\documentclass[conference]{IEEEtran}
\IEEEoverridecommandlockouts
\usepackage{cite}
\usepackage{amsmath,amssymb,amsfonts}
\usepackage{algorithm}
\usepackage{algorithmic}
\usepackage{graphicx}
\usepackage{subfigure}
\usepackage{textcomp}
\usepackage{xcolor}
\newtheorem{definition}{\bf Definition}[section]
\newtheorem{thm}{\bf Theorem}
\newtheorem{proof}{\bf Proof}
\def\BibTeX{{\rm B\kern-.05em{\sc i\kern-.025em b}\kern-.08em
    T\kern-.1667em\lower.7ex\hbox{E}\kern-.125emX}}
    
\begin{document}
\title{A Bionic Natural Language Parser Equivalent to a Pushdown Automaton\\
}
\author{
\IEEEauthorblockN{1\textsuperscript{st} Zhenghao Wei}
\IEEEauthorblockA{\textit{School of Computer Science} \\ \textit{and Engineering,} \\
\textit{Sun Yat-Sen University}\\
Guangzhou, China \\
0000-0003-3316-3589}
\and
\IEEEauthorblockN{2\textsuperscript{nd} Lin Kehua}
\IEEEauthorblockA{\textit{School of Computer Science}\\ \textit{and Engineering, }\\
\textit{Sun Yat-Sen University}\\
Guangzhou, China \\
linkh8@mail2.sysu.edu.cn}
\and
\IEEEauthorblockN{3\textsuperscript{rd} Jianlin Feng{*}}
\IEEEauthorblockA{\textit{School of Computer Science}\\ \textit{and Engineering,} \\
\textit{Sun Yat-Sen University}\\
Guangzhou, China \\
fengjlin@mail.sysu.edu.cn }
}




\maketitle

\begin{abstract}
Assembly Calculus (AC), proposed by Papadimitriou et al., aims to reproduce advanced cognitive functions through simulating neural activities, with several applications based on AC having been developed, including a natural language parser proposed by Mitropolsky et al. However, this parser lacks the ability to handle Kleene closures, preventing it from parsing all regular languages and rendering it weaker than Finite Automata (FA). In this paper, we propose a new bionic natural language parser (BNLP) based on AC and integrates two new biologically rational structures, Recurrent Circuit and Stack Circuit which are inspired by RNN and short-term memory mechanism. In contrast to the original parser, the BNLP can fully handle all regular languages and Dyck languages. Therefore, leveraging the Chomsky-Schützenberger theorem, the BNLP which can parse all Context-Free Languages can be constructed. We also formally prove that for any PDA, a Parser Automaton corresponding to BNLP can always be formed, ensuring that BNLP has a description ability equal to that of PDA and addressing the deficiencies of the original parser.
\end{abstract}


\section{Introduction}

Understanding human language comprehension has long been a central pursuitin linguistics, with extensive research in psycholinguistics, computational linguistics, and neuroscience offering numerous perspectives on the mechanisms underlying the human brain's understanding of language. Despite these advances, a significant gap remains between neurons and cognition. Recently, Papadimitriou et al. \cite{papadimitriou2020brain} introduced Assembly Calculus (AC), a brain computational system with high fidelity in both cognitive and biological aspects, offering the potential to simulate advanced human cognitive functions based on neural activities.

Drawing upon AC and Friederici's research on language in the brain \cite{LanguageInBrain}, Mitropolsky et al. \cite{mitropolsky2021biologically} proposed a bionic parser capable of parsing reasonably complex sentences in a biologically plausible manner. However, our experiments and their discussion reveal that this original parser (OP) struggles with sentences involving Kleene closures, such as multiple adjectives modifying a noun. To be considered a comprehensive parser for natural languages, it must at least be capable of handling all Context-Free Languages (CFLs), rendering the OP insufficient.

In this paper, we propose an enhanced {\itshape bionic natural language parser} (BNLP) based on AC, incorporating a Recurrent Circuit (RC) design inspired by the human brain's short-term memory mechanism. The RC can be intuitively thought of as the "Recurrent Neural Network (RNN)" of the AC system, extending the capabilities of the current feedforward normal form to express hierarchical structures by introducing a recurrent normal form that enables associations between an indefinite number of homogeneous objects. The RC also serves as a crucial component for maintaining excitement when the previous input stimulus declines or disappears, refers to Kar et al. \cite{evidenceofrc}.

We formally prove that our BNLP possesses a stronger descriptive ability than Finite Automata (FA), demonstrating that a {\itshape parser automaton} (PA) can be transformed from the BNLP. We establish that, for any FA, a PA accepting the language of the FA can always be constructed, allowing BNLP to handle all regular languages and rectify the OP's deficiencies. Moreover, according to the Chomsky-Sch{\"u}tzenberger theorem \cite{CHOMSKY1963118}, by proving that BNLP can parse both regular languages and Dyck languages, we establish that BNLP can indeed parse all CFLs. This finding addresses the concerns regarding the rigor of Mitropolsky et al.'s proof \cite{mitropolsky-etal-2022-center} and further supports the use of the BNLP as a parser for CFLs. That means, BNLP is equivalent to a pushdown automaton.


\section{THE BIONIC NATURAL LANGUAGE PARSER With Existing Problems}
\label{sec:2}
\subsection{Assembly Calculus}
Assembly Calculus ({\bf AC}) is a computational framework based on a human brain recently proposed by Papadimitriou et al. \cite{papadimitriou2020brain,DBLP:conf/innovations/PapadimitriouV19,papadimitriou2022bridging}, and can be regarded as an implementation of Buzsáki’s assembly-centered cognitive neuroscience theory \cite{buzsaki2010neural}. AC was created to answer the question of Richard Axel \cite{20181110}: ``we do not have a logic for the transformation of neural activity into thought and action.''. Papadimitriou et al. hope to establish a bottom-up brain computing model that uses the simulation of the underlying neural activity to reproduce the advanced cognitive functions of the brain. There are a series of operations in AC, which is "a formal system with a repertoire of rather sophisticated operations''.

We will briefly introduce the model of AC and the most critical operation, Random Projection and Cap (RP\&C), which is the basis of BNLP's operation mechanism. The brain is considered to be composed of a finite number $a$ of brain areas $A,B,C,\cdots$, which are regarded as Erd\"{o}s-R\'{e}nyi random graphs ($G(n,p)$, while there are $n$ neurons in an area and each pair of neurons has a probability $p$ of being connected by a synapse). The synapse $(i,j)$ connecting neuron $i$, $j$ has a non-negative weight $w_{ij}$ which represents the strength of the synapse and is initialized to 1. Besides neurons, brain areas are also connected by nerve fibers composed of clusters of synapses. For a pair of areas $(A,B)$, there is a fiber between them means that there is a random directed bipartite graph between neurons of $A$ and neurons of $B$, while the probability of there being synapse between a neuron in $A$ and a neuron in $B$ is also $p$. 

Then what needs to be introduced is the projection mechanism of AC. To think abstractly, time is divided into discrete steps, and in each step for each area where projection occurs, there are only $k$ neurons active. Those neurons activated are called ``{\itshape cap}'' or winners, and they have the highest synaptic inputs. The synaptic input of a neuron is the sum of the weights of synapses connected to this neuron whose pre-synaptic neuron was activated in the last step. The core mechanism of AC is the simulation of Hebbian plasticity through the process of projections. The idea of Hebbian plasticity summarized as ``cells that fire together wire together'' 
\cite{Shatz1992TheDB}, that means neurons always activated together will have higher synaptic weights. For a synapse whose pre-synaptic neuron and post-synaptic neuron fire successively, its weight will be enhanced by $(1+\beta)$, $\beta$ denotes the plasticity factor. 

The events happen from time $t$ to time $t+1$ can be described as below:

\begin{enumerate}
    \item Given the state of time $t$. The input of each neuron $i$ can be calculated by
    \begin{equation}
        {I_{i}}^t=\Sigma_{(j,i)\in E,{f_j}^t=1} w_{ji}^t
    \end{equation}
    Where ${I_{i}}^t$ denotes the input of $i$ in time $t$, $E$ is the set of synapses, ${f_j}^t=1$ means that neuron $j$ activated in time $t$, and $w_{ji}^t$ denotes the weight of $(j,i)$ at the present.
    \item Solve the top-k problem, to decide the $k$ neurons fired in the next step, by comparing ${I_{i}}^t$. For a winner $i$, it will have a ${f_i}^{t+1}=1$.
    \item Update the weights. For each synapse $(i,j)\in E$, will do:
    \begin{equation}
        {w_{ij}}^{t+1}={w_{ij}}^t (1+f_i^t f_j^{t+1} \beta)
    \end{equation}
\end{enumerate}

Through this projection process, the densely intraconnected assemblies as the same as D.O.Hebb predicted will be formed \cite{Hebb1949TheOO}, and they can stably represent objects in the brain like concepts, words, figures, etc. Furthermore, an assembly can be connected with several assemblies from different areas, to form a graph structure, which can represent complex hierarchical structure, like the parse tree.

\subsection{Biological Plausible Parser}
\label{2b}
 Papadimitriou el al. chose Natural Language Understanding as the killer application of AC, mainly because the language ability is the most fundamental and prominent feature of the human brain distinguishing from animals \cite{10.7551/mitpress/11173.001.0001}. Moreover, the design of AC is greatly inspired by Chomsky's formal language theory, e.g. the Merge operation is derived from Chomsky normal form. A framework utilizing AC to parse language was proposed in \cite{papadimitriou2020brain}, which uses Merge operation to build a hierarchical parse tree. Intuitively, all context-free grammars that can be represented by Chomsky normal form (CNF) can be parsed. However, this idea was not adopted by the follow-up work including the original parser.
 
The original parser, proposed by Mitropolsky et al. \cite{mitropolsky2021biologically} is the first attempt to achieve the goal of reproducing human language ability through AC. Based on Friederici et al.'s research on language organs \cite{LanguageInBrain,10.3389/fpsyg.2015.01818}, it, to some extent, reproduces the process of human brain understanding language from vocabulary recognition to syntax parsing. The process starts with an access to a {\itshape lexicon} area, which is thought to reside in the left medial temporal lobe (MTL) and is represented as a look-up table of vocabulary. Then, in the Wernicke’s area in the superior temporal gyrus (STG), the syntactic roles of words are identified.

In principle, OP abandons the operation-based system in \cite{papadimitriou2020brain}. Instead, it is based on the theory of language organs, and believes that the brain areas representing different syntactic roles have corresponding apriori area rules and fiber rules. Based on these rules, in the process of understanding language, the map of the brain's next projection will be automatically generated.

The experiment in Mitropolsky et al.'s work \cite{mitropolsky2021biologically} shows that OP can parse many reasonably non-trivial sentences at a high frequency range of 0.2-0.5 seconds/word. However, they did not formally prove the boundary of OP's parsing ability. Actually, OP cannot resolve the case where unknown multiple consecutive adjectives modify a noun. In other words, it cannot handle Kleene closure, and therefore it cannot handle all syntactic categories. Although two imperfect improvement schemes were mentioned in \cite{mitropolsky2021biologically}, neither can actually handle Kleene closure. 

\begin{figure}[htbp]
    \centering
    \includegraphics[width=\linewidth]{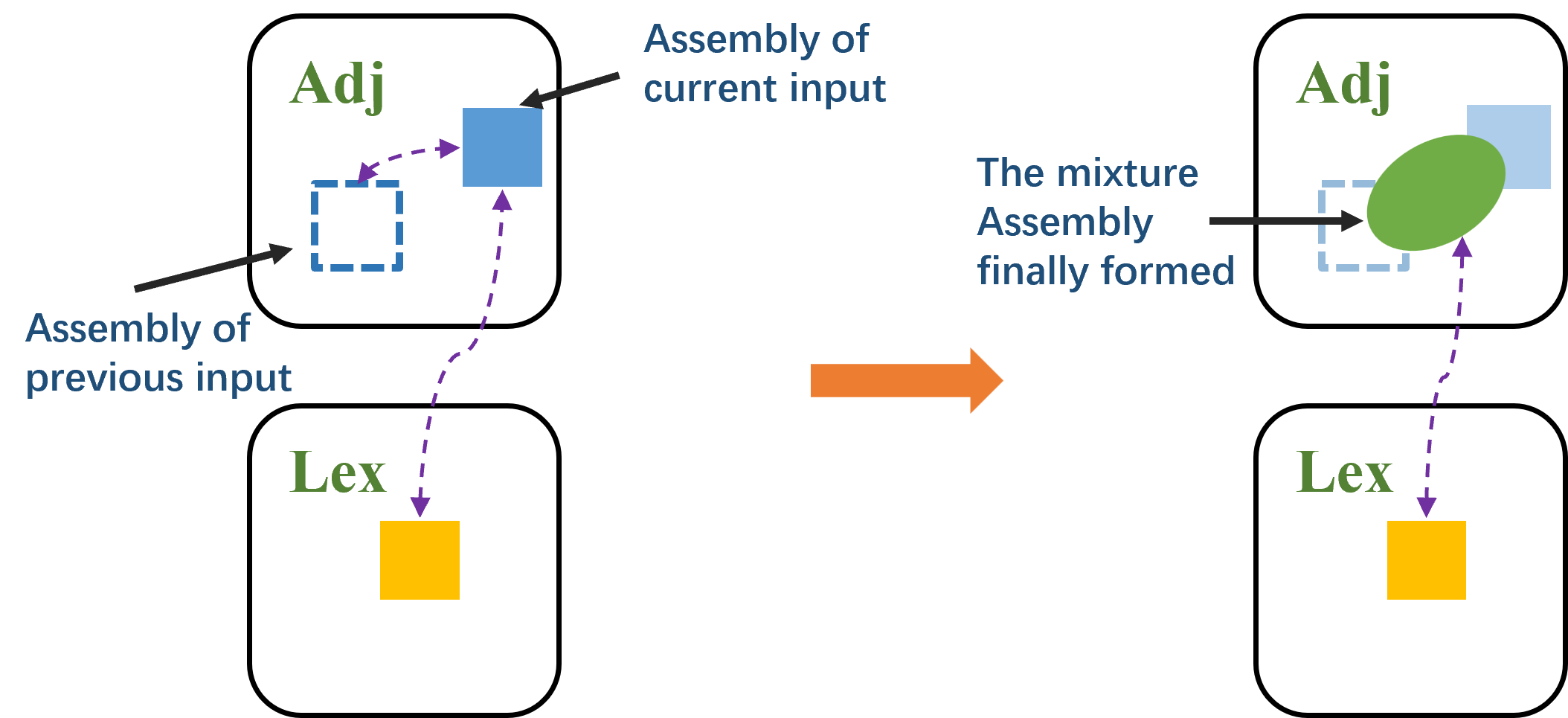}
    \caption{When OP accepts multiple consecutive adjectives, there will be confusion among the neurons in the assemblies of two adjacent adjectives.}
    \label{fig:mix}
\end{figure}

The main reason why OP cannot handle Kleene closure is that, in the design of OP, when multiple consecutive words with the same part-of-speech appear, multiple stable assemblies cannot be activated simultaneously in the same brain area, or there will be confusion among their neurons as shown in Figure \ref{fig:mix}. In spite of the deficiency of OP, the
language understanding models based on AC still seem to have great potential, considering that Papadimitriou et al. have proved in \cite{papadimitriou2020brain} that AC has the computing power equivalent to that of a Turing machine, which means OP should possess a more powerful parsing capability under appropriate design.

It is worth mentioning that Mitropolsky et al. recently proposed a parser variant \cite{mitropolsky-etal-2022-center}, which contains a new operation called ``{\itshape fallback-again}'' to deal with the case of Center-Embedding (e.g. clauses). It is claimed that the parser variant could deal with Context-Free Language ({\bf CFL}), however, the proof was just an outline and was based on unreliable assumptions, in which there are several flaws:

\begin{enumerate}
    \item In the proof, OP had to have the ability to parse RE as a presupposition. As has been pointed out, if OP cannot resolve the cases of multiple adjectives modifying a noun, then OP's expressive ability will be less than that of regular language (RL).
    \item Mitropolsky et al. misunderstood the classic theorem of Chomsky \cite{CHOMSKY1963118}. They stated a CFL $L$ can be written in the form of $L=R\cap h(D_K)$, but the correct form is $L=h(R\cap D_K)$, where $R$ represents a RL, $D_K$ denotes a Dyck language and $h$ signifies a homomorphism. The wrong form indicates that CFL should be a subset of RL, which is unrealistic.
    \item Actually, the Fallback-again operation is not truly capable of handling center-embedding, and it suffers from several issues. For example, when two(or more) identical leading words appear consecutively, the DS area cannot differentiate between the two(or more) words, resulting in a failure of syntactic parsing. In summary, the ability of this solution to parse Dyck languages is questionable and cannot be taken for granted in the development of BNLP.
\end{enumerate}

Therefore in this study, we develop our BNLP and fill in the deficiencies of Mitropolsky et al’s OP. As a result, BNLP can represent RLs and Dyck languages, and even deal with all CFLs.

\section{Methods}
\label{sec:3}
\subsection{The original parser}
The main data structure of OP is an undirected graph $G=(\mathcal{A},\mathcal{F})$, where $\mathcal{A}$ is the set of brain areas and $\mathcal{F}$ is the set of fibers connecting areas. There is an essential area {\itshape Lexicon} (or {\itshape Lex} in short) representing a look-up table of words in left MTL, and the other brain areas represent different parts-of-speech or syntactic components, like Verb, Sbj, Obj, etc. 

All areas except {\itshape Lex} are called normal areas of AC, each of which contains $n$ neurons and the limit of winners is $k$ \footnote{In the experiments, usually, $n=10^6, p=10^{-3}, k=10^3$, or for fast calculation $n=10^4,p=0.01,k=10^2$.}.The Lex area is called the {\itshape explicit area}, in which a fixed set of neurons represent a word and explicitly stored in a word list. When a word is coming into the parser, the corresponding neurons in Lex will be directly fired instead of producing a random stimulus. These neurons are called fixed assemblies, and the assembly corresponding to word $w$ is named as $x_w$.

In the non-lexicon areas, the mechanisms to control the projection process to practice those innate or acquired grammatical knowledge is abstracted as {\itshape fiber-rules} or {\itshape area-rules}.  They are collectively called {\itshape actions}. The action $\alpha_w$ of a word $w$ is related to its part-of-speech, and consists of two sets of commands which are called {\itshape pre-rules} and {\itshape post-rules}. The pre-rules are applied before the current projection and the post-rules are applied after the current projection. There are only two kinds of rules in the actions, {\itshape disinhibit} and {\itshape inhibit}, and the targets of inhibition or disinhibition can be areas or fibers. By default, all areas and fibers are inhibited and do not participate in any projection unless they are disinhibited. In the real human brain, there also exists a phenomenon called {\itshape symmetry breaking}, that is, only a few neurons and areas are frequently activated, while the vast majority of neurons are suppressed at most times \cite{Jphy22,AGexciteality}. Only when an area itself is disinhibited and a fiber pointing to it from an area with activated assembly is disinhibited, that area can participate in the next projection.

Since OP abandons the operation-based AC system, the projection operation in OP needs to be specially defined, which actually consists of a series of RP\&C operations being executed at once. Mitropolsky et al. denotes the operation of executing a set of projections at once as {\itshape strong projection} or {\itshape Projection*}. We denote the set of projections being executed as {\itshape projection map}, for they are stored as a map in the program. Whenever a pre/post-rule is applied, there might be a new projection being added to the map, as defined in definition \ref{projmap}.

\begin{definition}[Projection map]
\label{projmap}
    A projection map is a set of projections to be executed in a Projection* operation. In the original AC system, there are two kinds of map, stimli-area map and area-area map. However, there is no stimulus in OP, so only area-area map is considered in this paper. An area-area map is a series of key-value pairs in the form of $ \{from\_area_1: [to\_area_1,to\_area_2,\cdots ], from\_area_2:[\cdots ], \cdots \} $, in which a key represents the from-area and the value represents a list of to-areas of the projection. 

    For two areas $A$ and $B$, only when $A$ has an assembly activated in the last step and both area $B$ and the fiber between $A$ and $B$ are disinhibited, the projection from $A$ to $B$ will be added to the map. The projection is denoted as $A\rightarrow B$. 
\end{definition}

\begin{figure*}[!htbp]
    \centering
    \begin{minipage}{\linewidth}
    \subfigure[]{
        \begin{minipage}{0.3\linewidth}
            \centering
            \includegraphics[width=\linewidth]{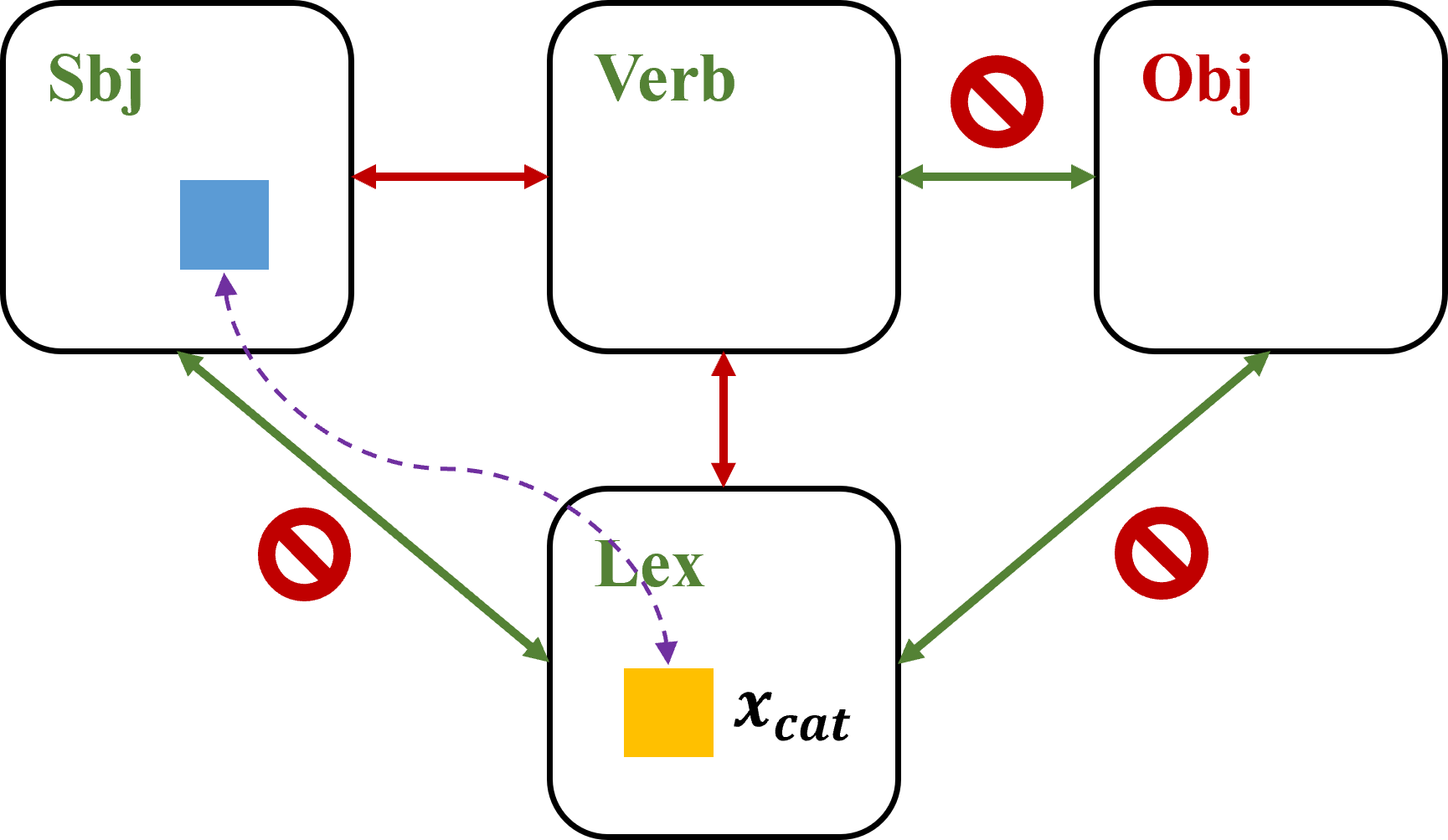}
        \end{minipage}
    }
    \subfigure[]{
        \begin{minipage}{0.3\linewidth}
            \centering
            \includegraphics[width=\linewidth]{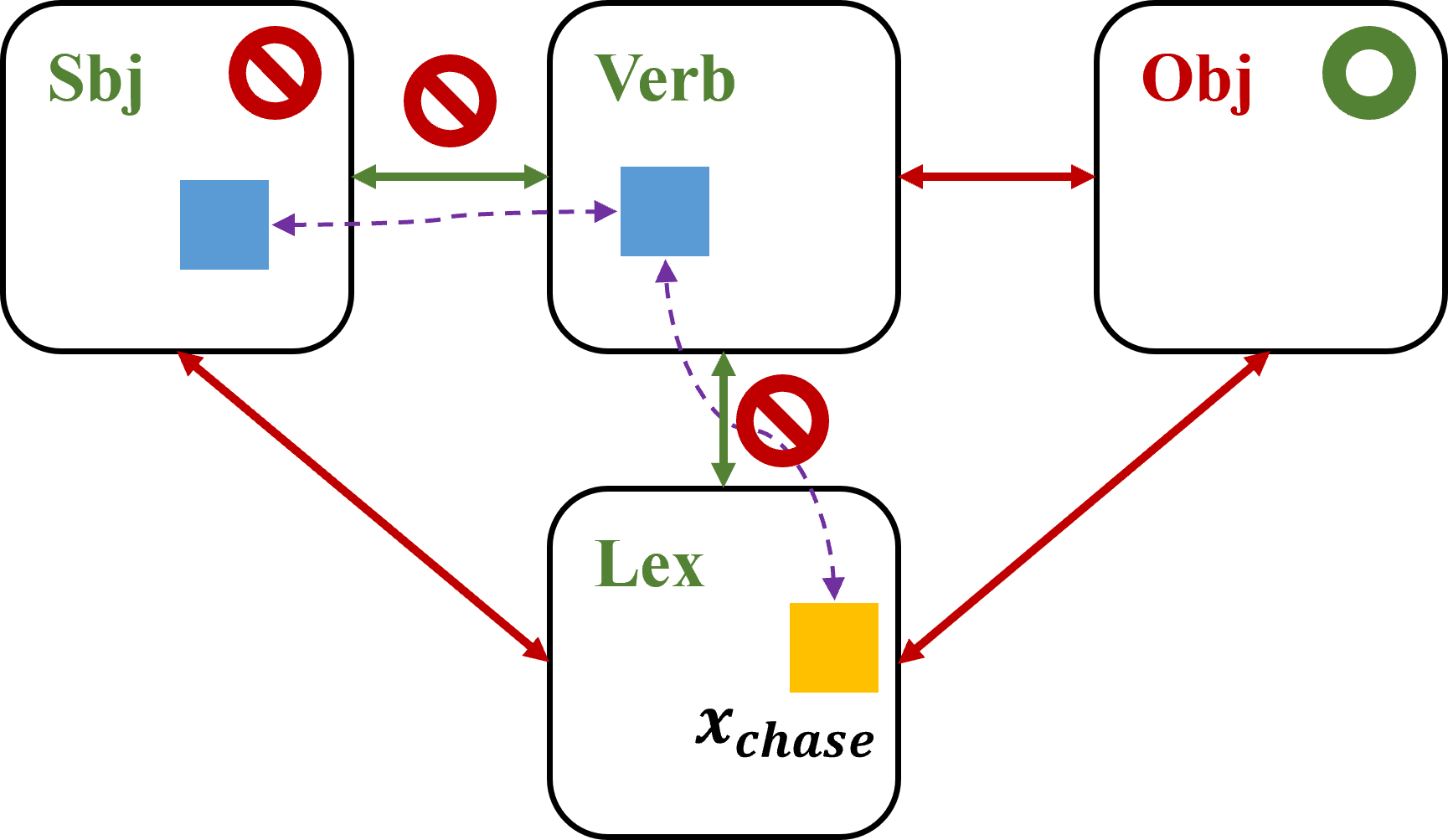}
        \end{minipage}
    }
    \subfigure[]{
        \begin{minipage}{0.3\linewidth}
            \centering
            \includegraphics[width=\linewidth]{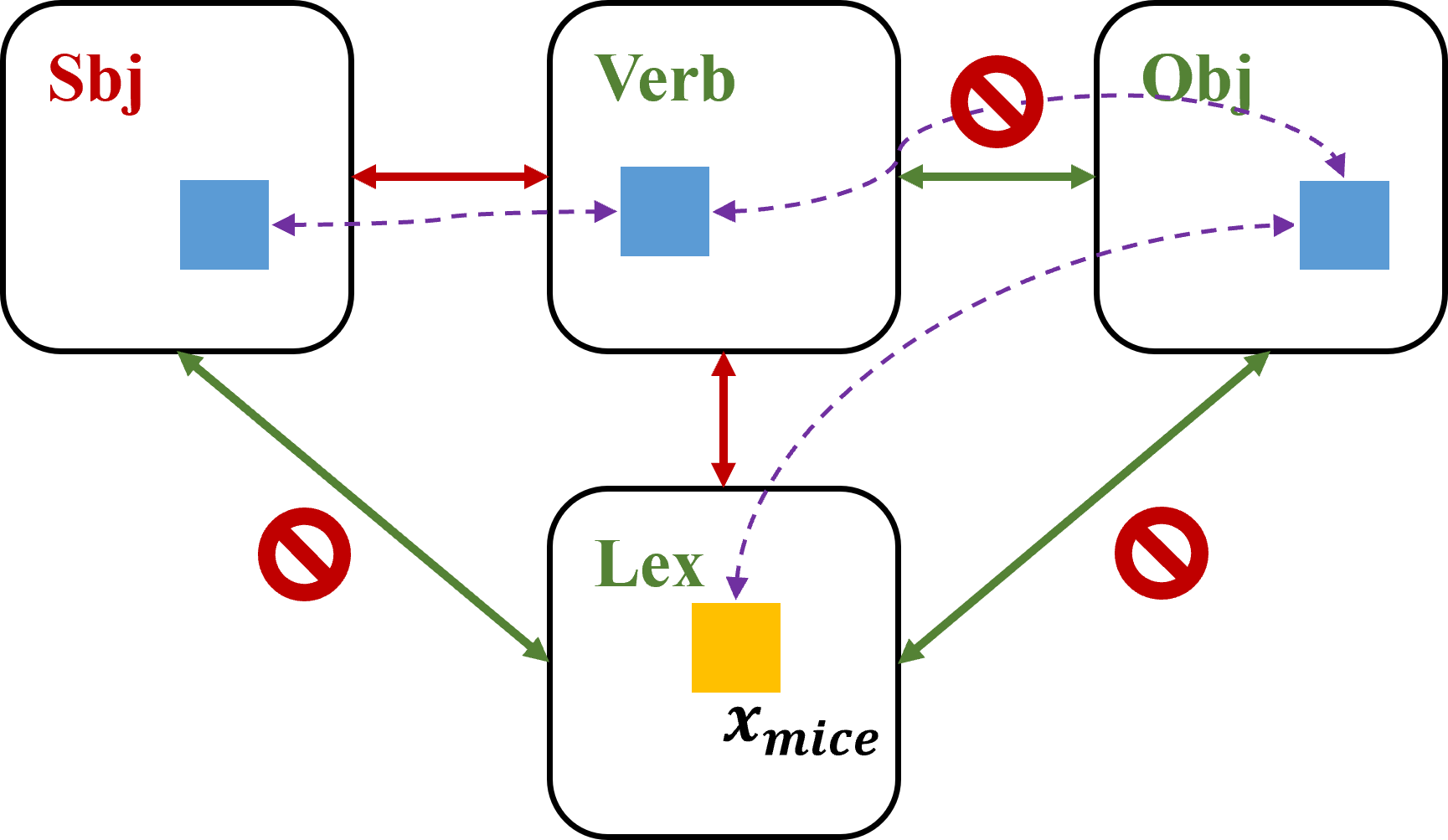}
        \end{minipage}
    }
    \end{minipage}
    \caption{An example of how OP operates, when the input sentence is "cat chase mice". Green areas and fibers are disinhibited and red ones are inhibited, {\color{red} {\bf $\oslash$}} means the area/fiber will be inhibited by the post-rules, and  {\color{green} {\bf $\bigcirc$}} means it will be disinhibited.}
    \label{fig:op}
\end{figure*}

The detailed description of OP is given in algorithm \ref{alg:algorithm1}\footnote{For convenience, (dis)inhibit($\square$,$\square$) is used to represent inhibit or disinhibit an area or a fiber, where $\square$ represents an area or ($\square$,$\square$) represents a fiber.} and Figure \ref{fig:op}. Let's take a look at a simple example ``cats chase mice'', with two nouns and one verb which respectively belong to three syntactic categories: Sbj, Verb, Obj. No matter a noun is a subject or an object, its corresponding action is consistent, i.e.,
\begin{small}
\begin{equation*}
    \alpha_{cat}=
    \left\{
    \begin{array}{c|c}
        pre\text{-}rules: &  post\text{-}rules:\\
        disinhibit(Lex,Sbj) & inhibit(Lex,Sbj)\\
        disinhibit(Lex,Obj) & inhibit(Lex,Obj)\\
        disinhibit(verb,Sbj) & inhibit(Verb,Obj)\\
    \end{array}
    \right\}
\end{equation*}
\end{small}
and the action of a verb is like:
\begin{small}
\begin{equation*}
    \alpha_{chase}=
    \left\{
    \begin{array}{c|c}
        pre\text{-}rules: &  post\text{-}rules:\\
        disinhibit(Lex,Verb) & inhibit(Lex,Verb)\\
        disinhibit(Sbj,Verb) & inhibit(Sbj,Verb)\\
        \  & inhibit(Sbj)\\
        \  & disinhibit(Obj)\\
    \end{array}
    \right\}
\end{equation*}
\end{small}

Moreover, when a verb is coming after a noun, a projection \{Lex:[Verb], Sbj:[Verb]\} is added to the projection map.

As the result of parsing, a cluster of assemblies is formed, with neurons connected by dense synapses with heavy weights. Then, the parse tree can be reproduced from the graph relationship among these assemblies.

\begin{algorithm}[ht] 
\renewcommand{\algorithmicrequire}{\textbf{Input:}}
\renewcommand{\algorithmicensure}{\textbf{Output:}}
\caption{Algorithm: main loop of OP} 
\label{alg:algorithm1} 
\begin{algorithmic}[1]
\REQUIRE a sentence $s$
\ENSURE a parse tree represents $s$, rooted in the predicate
\STATE disinhibit(Lex);
\STATE disinhibit(Sbj);
\STATE disinhibit(Verb);
\FORALL{word $w$ in $s$}
    \STATE activate assembly $x_w$ in Lex;
    \FORALL{pre-rule (dis)inhibit($\square$,$\square$) in $\alpha_w\rightarrow{pre\text{-} rule}$}
    \STATE (dis)inhibit($\square$,$\square$);
    \ENDFOR
    \STATE Produce the projection map and perform Projection*;
    \FORALL{post-rule (dis)inhibit($\square$,$\square$) in $\alpha_w\rightarrow{post\text{-} rule}$}
    \STATE (dis)inhibit($\square$,$\square$);
    \ENDFOR
\ENDFOR
\STATE readout();
\end{algorithmic}
\end{algorithm}

According to the studies in \cite{mitropolsky2021biologically}, OP can only parse the natural language sentences whose structures follow 20 particular templates. It cannot handle phrases like ``a big fat bad orange cat'', in which there are unknown multiple consecutive adjectives or adverbs. Specifically, it cannot handle the case of Kleene closure, and thus it fails to parse all syntactic categories. This limitation of OP motivates us to develop a parser with recurrent circuits.

\subsection{Memory and Recurrent Circuits in the brain}
Recurrent Neural Network (RNN) was first described by Jordan in 1986 (Jordan Network) \cite{JORDAN1997471} and then reported by Elman \cite{ELMAN1990179}. Nowadays RNN and its variants have been widely used in Natural Language Processing (NLP) and become a vital tool for deep learning \cite{DBLP:conf/interspeech/MikolovKBCK10,LSTM,DBLP:journals/corr/ChungGCB14,encoder}. Although in recent years RNN is mainly used as a sequential feature extractor to extract features with temporal or spatial dependencies from various scenes, RNN was originally proposed as a neural network with memory. 

RNN is an abstraction of structures existing in the nervous system with solid anatomical basis ({\bf reverberating circuit} or {\bf recurrent circuit}) \cite{principleofconnex}, although there are different views on the interpretation of the cognitive function of these structures. Generally speaking, there are two basic normal forms in a Hebbian brain model: one is {\bf feedforward circuit}, the other is {\bf recurrent circuit} ({\bf RC}). The hierarchical feedforward circuit can describe the process of nerve impulse transmission from the stimuli to functional brain areas, while the recurrent circuit describes the process of how non-exogenous excitatory signals are maintained in the nervous system, and is also considered as the basis for the existence of short-term memory \cite{LitwinKumar2014FormationAM,evidenceofrc}. 

In the original AC version in \cite{papadimitriou2020brain}, the feedforward case has been fully discussed, just as building a graph of assemblies to store a parse tree, the clustering of assemblies are indeed a feedforward network. But the recurrent circuit is still not included. We believe that the introduction of RC will greatly complement the limitations of AC, especially for those scenarios that cannot be covered by neurons with certain number of layers.



\subsection{Parser with recurrent circuits}
\begin{figure}[H]
\centering
    \begin{minipage}[t]{0.8\linewidth}
        \centering
        \includegraphics[width=\textwidth]{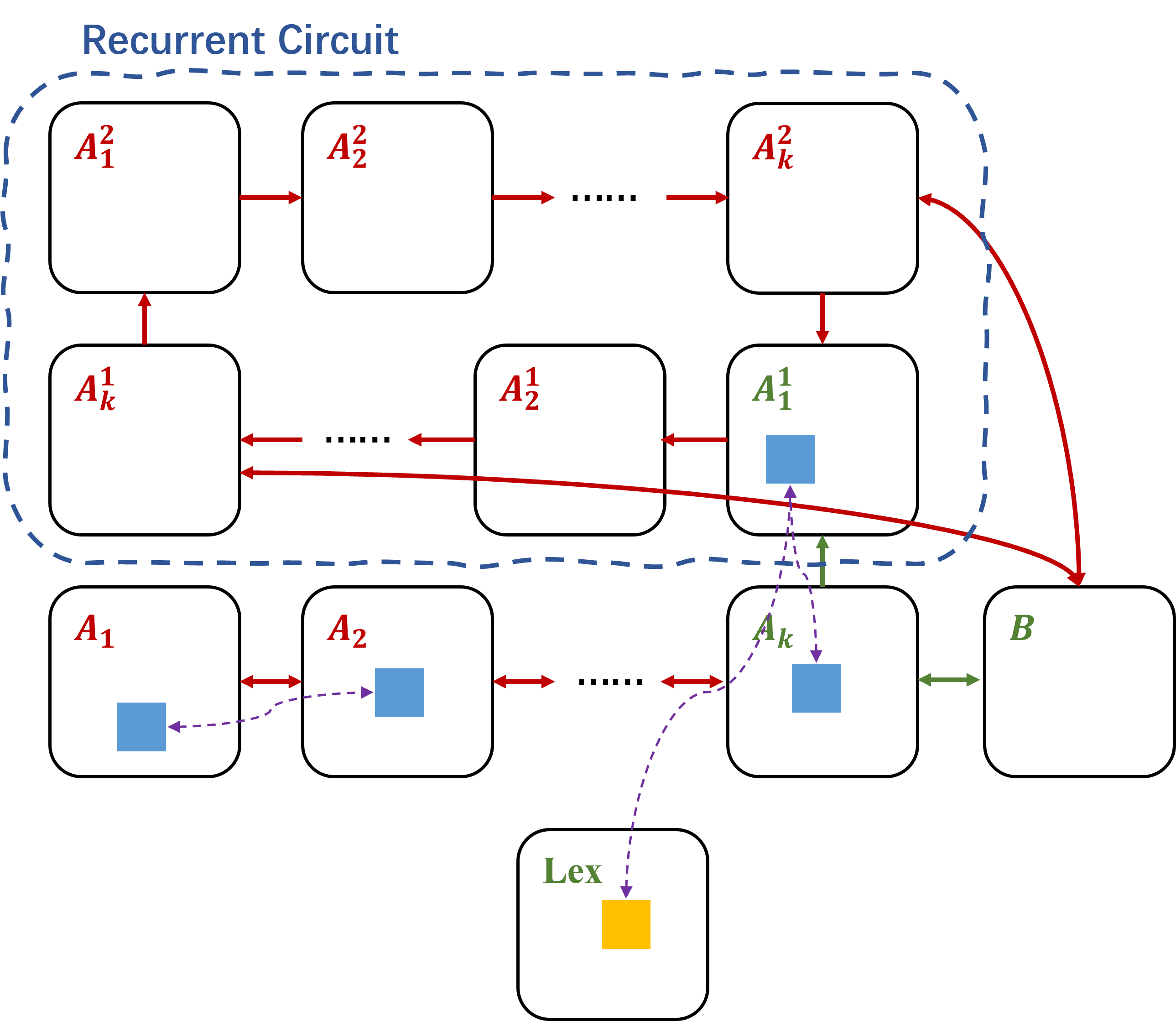}
        \label{rc1}
    \end{minipage}%
    \caption{A sketch of RC. To make the figure clear, fibers connecting Lex and other areas are omitted. Fibers inside RC are all directed fibers, and fibers connected RC areas with other areas are undirected.}
    \label{fig3}
\end{figure}

A Recurrent Circuit is able to deal with a group of homogeneous objects with uncertain length in a circuit with certain scale. We import it to solve the case of Kleene closure in natural language parsing. While it is obviously impossible for the brain to always prepare enough distinct areas to cope with uncertain external stimuli, RC can act as a ``stimulus buffer'' to deal with such a case. The structure of an RC can be regarded as a series of areas connected by {\itshape directed} fibers, forming a strongly connected graph, in which each area has only one outgoing edge and one incoming edge. Figuratively speaking, this graph is like a ``ring'', as shown in Figure \ref{fig3}.

Note that fibers in OP are undirected edges. Thus, the definition of {\itshape fiber} should be updated as follows.

\begin{definition}[Fiber]
\label{def:fiber}
There are two kinds of fibers. The default one is {\itshape undirected fiber} (simply called {\itshape fiber}) denoted by $\overline{AB}$, and the other is {\itshape directed fiber} denoted by $\overrightarrow{AB}$. 

When we say that there is a {\itshape fiber} $\overline{AB}$ between areas $A$ and $B$, it means the neurons in $A$ and $B$ form a bipartite graph in which the probability of a neuron in $A$ and a neuron in $B$ having a synapse is $p$, and vice versa. If the {\itshape fiber} $\overline{AB}$ is disinhibited during the projection process, both $A\rightarrow B$ and $B\rightarrow A$ should be added to the projection map. 

When we say that there is a {\itshape directed fiber} $\overrightarrow{AB}$, it means the probability of a neuron in $A$ having a synapse leading to any neuron in $B$ is $p$. If the {\itshape directed fiber} $\overrightarrow{AB}$ is disinhibited during the projection process, only $A\rightarrow B$ should be added to the projection map. $\square$

\end{definition}

Next we introduce the specific design of RC in BNLP.

\begin{definition}[Recurrent Circuit]
  A Recurrent Circuit is a group of brain areas connected {\itshape end-to-end} by directed fibers. {\itshape End-to-end} connection means that these areas are arranged in a sequence, where for any two adjacent areas $A_i$ and $A_{i+1}$ in the arrangement, there is a directed fiber $\overrightarrow{A_iA_{i+1}}$. From the perspective of graph theory, an RC is a strongly connected graph in which each vertex has only one outgoing edge and one incoming edge. 

For a parsable Kleene closure $(R)^*$, $R$ is a regular expression that can be parsed with $k$ non-lexicon areas $\{A_1,A_2,\cdots,A_k\}$ \footnote{If there are no more Kleene stars in the closure, then $k$ is exactly the number of symbols in the closure.}. In RC, $(R)^*$ is parsed with $2k$ areas, which are exactly two copies of the above $k$ areas namely $\{A_1^1,A_2^1,\cdots,A_k^1\}$ and $\{A_1^2,A_2^2,\cdots,A_k^2\}$ \footnote{The reason why two copies are needed is to avoid the situation mentioned in Figure \ref{fig:mix}, e.g., if there is only one symbol in the closure and only one copy is made, RC will only contain one area so that the confusion phenomenon among assemblies will still exist.}. So the number of areas in RC must be even.

For the two copies of area $A_i$, i.e., $A_i^1$ and $A_i^2$, they have the same actions as $A_i$ based on the syntactic categories. In addition, the pre-rules and post-rules for each area $A_i^j$ need to be added, i.e.,

    \begin{equation}
    \left\{
    \begin{array}{c|c}
        pre\text{-}rules: &  post\text{-}rules:\\
        disinhibit(A_{i-1}^j,A_i^j) & inhibit(A_{i-1}^j,A_i^j)\\
        disinhibit(Lex,A_i^j) & inhibit(Lex,A_i^j)\\
    \end{array}
    \right\}
    \label{eq:3}
    \end{equation}

    where $j\in \{1,2\}$. In the rules of $A_1^j$, the area pair should be replaced by $(A_{k}^{3-j},A_1^j)$ as shown in Figure \ref{fig3}. Equation (\ref{eq:3}) is mainly for the case of having only one symbol in the closure, because when $A_1^j$ is to be projected, fiber $(Lex,A_{k}^{3-j})$ has already been inhibited.
    
    Besides the fibers inside the RC areas, the fibers connecting the RC areas and other areas also need to be considered. For an RC $\{ A_1^1,A_2^1,\cdots,A_k^1,A_1^2,A_2^2,\cdots,A_k^2 \}$ with the original areas $\{A_1,A_2,\cdots,A_k\}$, if there exists an area $B$ that follows the last area of the closure $A_k$, the following directed fibers must be added: 
    \begin{equation*}
        \overrightarrow{A_kA_1^1},\overline{A_k^1B},\overline{A_k^2B}
    \end{equation*}

    In addition, several fiber-rules also need to be added for $A_k^j$:

    \begin{equation}
    \left\{
    \begin{array}{c|c}
        pre\text{-}rules: &  post\text{-}rules:\\
        & disinhibit(A_k^j,B)\\
        & inhibit(A_k^{3-j},B)\\
    \end{array}
    \right\}
    \label{eq:4}
    \end{equation}

    At the same time, the rules of $A_1^1$ must be updated to get rid of the impact of $A_k$:
    \begin{equation}
    \left\{
    \begin{array}{c|c}
        pre\text{-}rules: &  post\text{-}rules:\\
        & inhibit(A_k,A_1^1)\\
        & inhibit(A_k,B)\\
        & inhibit(A_k)\\
    \end{array}
    \right\}
    \label{eq:5}
    \end{equation}
    
\end{definition}

Figure \ref{fig3} illustrates the establishment of the RC. Based on these definitions, the fibers connecting areas inside and outside the RC can be easily written, as well as the action of each area. 

\begin{figure}[htbp]
\centering
    \begin{minipage}[t]{0.9\linewidth}
    \subfigure[Process the 1\textsuperscript{st} and 2\textsuperscript{nd} adjectives]{
        \centering
        \includegraphics[width=\linewidth]{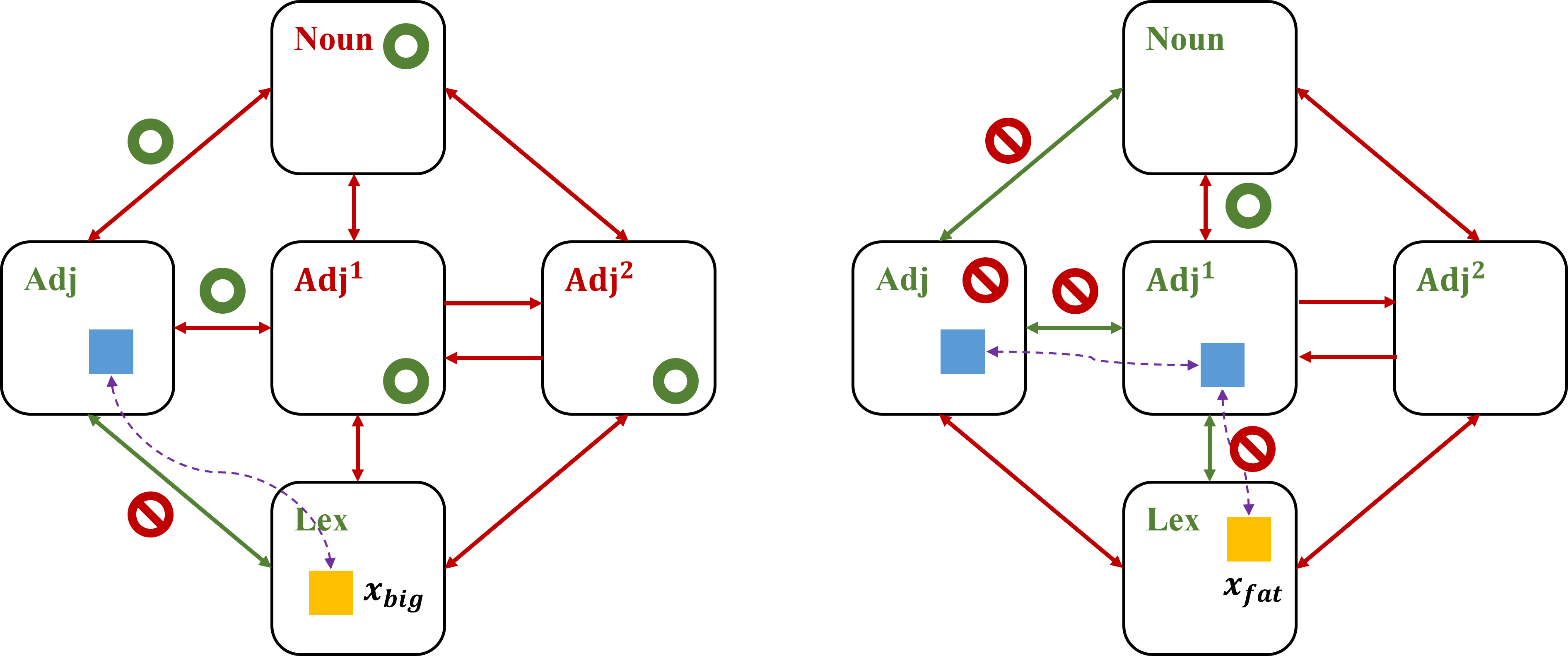}}
    \end{minipage}%
    
    \begin{minipage}[t]{0.9\linewidth}
    \subfigure[Process the 3\textsuperscript{rd} and 4\textsuperscript{th} adjectives]{
        \centering
        \includegraphics[width=\linewidth]{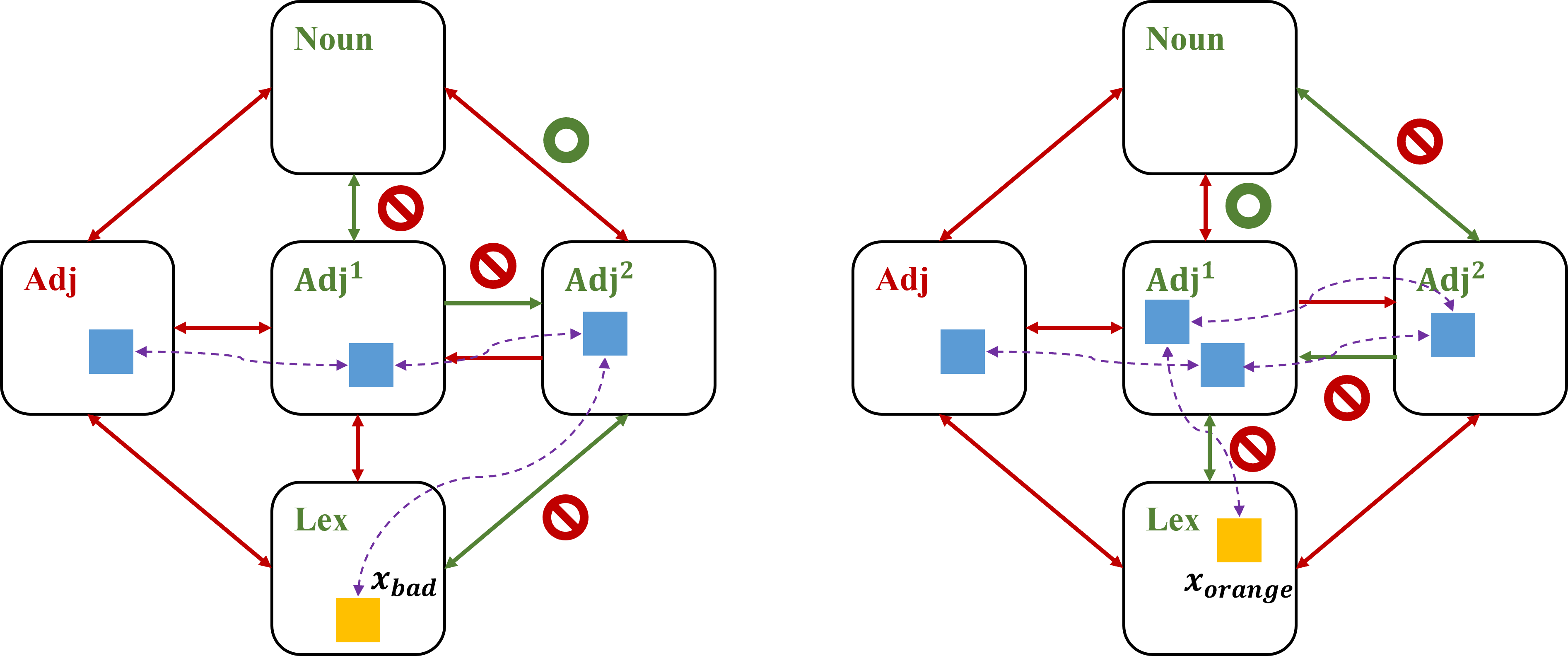}}
    \end{minipage}
    \caption{An example of inputting four consecutive adjectives while parses a string conforms to ``$(Adj)^*\ N$'', where $Adj$ accepts any adjective. In the figure, {\itshape $Adj^1$} and {\itshape $Adj^2$} compose a minimum RC. Through ``tossing about'' projection, the relationship of a series of adjectives modifying a noun can be stored in RC.}
    \label{fig:rce}
\end{figure}

In Figure \ref{fig:rce}, we show how to utilize RC to parse a phrase with a series of adjectives modifying a noun, i.e., ``a big fat bad orange cat''. In this case, two Adj areas $Adj^1$ and $Adj^2$ need to be included in a minimum RC, both of which are the copies of the area $Adj$. The first adjective input is handled by the $Adj$ area outside RC, and the 2\textsuperscript{nd} to the 4\textsuperscript{th} adjective are processed by RC. In Figure \ref{fig:rce}, we explain how to parse the example in detail and show the construction of the projection map step by step. It is worth pointing out that when the 4\textsuperscript{th} adjective assembly is formed on $Adj^1$, for there is no neuronal excitation on $Adj^1$ at the previous moment, so confusion does not occur like in Figure\ref{fig:mix}, and different assembly can be successfully formed.

When the first adjective ``big'' is inputted, an action of adjective is applied, as:
\begin{small}
    \begin{equation*}
    \left\{
    \begin{array}{c|c}
        pre\text{-}rules:&  post\text{-}rules:\\
        disinhibit(Lex,Adj) & inhibit(Lex,Adj)\\
        disinhibit(\square,Adj) & inhibit(\square,Adj)\\
        & disinhibit(Adj,Noun)\\
        & disinhibit(Adj^1)\\
        & disinhibit(Adj^2)\\
        & disinhibit(Noun)\\
        & disinhibit(Adj,Adj^1)\\
    \end{array}
    \right\}
\end{equation*}
\end{small}

Then the projection map of adjective ``big'' is \{$Lex$:[$Adj$], $\square$:[$Adj$]\}, with the area before Adj denoted as $\square$. Next, the action of the second adjective, which is accepted by $Adj^1$, is as follows:
\begin{small}
\begin{equation}
\begin{split}
    \left\{
    \begin{array}{c|c}
        pre\text{-}rules: &  post\text{-}rules:\\
        disinhibit(Lex,Adj^1) & inhibit(Lex,Adj^1)\\
        disinhibit(Adj^2,Adj^1)& inhibit(Adj^2,Adj^1)\\
        & inhibit(Adj)\\
        & inhibit(Adj,Adj^1)\\
        & inhibit(Adj,Noun)\\
        & disinhibit(Adj^1,Noun)\\
        & inhibit(Adj^2,Noun)\\
    \end{array}
    \right\}
\end{split}
\label{Adj^1}
\end{equation}
\end{small}

The projection map becomes \{$Lex$:[$Adj^1$], $Adj$:[$Adj^1$]\} now. Notice that equation (\ref{Adj^1}) is a combination of equations (\ref{eq:3}), (\ref{eq:4}) and (\ref{eq:5}), because the area $Adj^1$ conforms to both $A_1^1$ and $A_k^j$. But for the 3\textsuperscript{rd} and 4\textsuperscript{th} adjectives, they will only conforms to $A_k^j$, so that the actions will be combination of equation (\ref{eq:3}) and (\ref{eq:4}), which are as follows:

\begin{small}
\begin{equation}
\begin{split}
    \left\{
    \begin{array}{c|c}
        pre\text{-}rules: &  post\text{-}rules:\\
        disinhibit(Lex,Adj^2) & inhibit(Lex,Adj^2)\\
        disinhibit(Adj^1,Adj^2)& inhibit(Adj^1,Adj^2)\\
        & disinhibit(Adj^2,Noun)\\
        & inhibit(Adj^1,Noun)\\
    \end{array}
    \right\}
\end{split}
\label{Adj^2}
\end{equation}
\end{small}

and

\begin{small}
\begin{equation}
\begin{split}
    \left\{
    \begin{array}{c|c}
        pre\text{-}rules: &  post\text{-}rules:\\
        disinhibit(Lex,Adj^1) & inhibit(Lex,Adj^1)\\
        disinhibit(Adj^2,Adj^1)& inhibit(Adj^2,Adj^1)\\
        & disinhibit(Adj^1,Noun)\\
        & inhibit(Adj^2,Noun)\\
    \end{array}
    \right\}
\end{split}
\label{Adj^11}
\end{equation}
\end{small}

The key-value pairs \{$Lex$:[$Adj^2$], $Adj^1$:[$Adj^2$]\} and \{$Lex$:[$Adj^1$], $Adj^2$:[$Adj^1$]\} are further added to the projection map. If there are more than four adjectives in the phrase to be parsed, the process shown in figure \ref{fig:rce}(b) repeats continuously, and the actions will be the same as Eq. (\ref{Adj^2}) and Eq. (\ref{Adj^11}).



\subsection{Center-embedding and Dyck language}
After establishing the Recurrent Circuit (RC), we will demonstrate in the following sections that our parser with RC has the ability to parse all RLs. With this capability in place, a more challenging question arises: how can we construct a parser that can parse all context-free languages? Fortunately, there is a useful theorem available to us, known as the Chomsky-Schützenberger (CS) theorem, which is also referred to as the CFL representation theorem.

There are various representations of the theorem, but in essence, the CS theorem states that {\itshape every context-free language can be represented as the intersection of a RL and a Dyck language under a homomorphism}. This theorem provides a powerful insight into the structure of context-free languages and serves as the foundation for constructing a parser that can handle all CFLs.

Thus, the only remaining task to prove that BNLP can parse context-free languages is to demonstrate that it can express Dyck languages in a way that is compatible with and does not conflict with the Recurrent Circuit (RC). If this can be achieved, BNLP will be able to represent the intersection of RLs and Dyck languages (DL). Similar to our analysis of BNLP's parsing capability for RLs, we must provide a strict definition of the operation of center-embedding and demonstrate whether or not it can effectively parse Dyck languages.

A Dyck language is a formal language that consists of well-balanced strings of pairs of opening and closing symbols or brackets. Mathematically, a Dyck language $D_k$ can be defined over an alphabet $\Sigma$, consisting of $k$ pairs of matching symbols (like brackets), denoted as $\{a_1, a_2, \cdots,a_k, \cdots, b_1, b_2, \cdots,b_k\}$.

The Dyck language $D_k$ is the smallest language satisfying the following conditions:

\begin{enumerate}
    \item The empty string $\epsilon$ is in $D_k$.
    \item If $x$ is in $D_k$ and $i \in \{1, 2, \cdots, k\}$, then $a_i x b_i$ is in $D_k$.
    \item If $x$ and $y$ are in $D_k$, then the concatenation of $x$ and $y$, $(xy)$ is in $D_k$.
\end{enumerate}

In other words, a string belongs to the Dyck language $D_k$ if and only if it is well-balanced with respect to the opening and closing symbols from the alphabet $\Sigma$. The simplest Dyck language is the balanced brackets language $D_1 = \{(,)\}$.

In order to cope with Dyck languages that feature an uncertain number of center-embeddings, the BNLP must incorporate a stack-like structure. We introduce a construct called the Stack Circuit (SC), which is somewhat similar to the RC. The SC can be understood as a sequence of areas arranged in a queue and connected by directed fibers. For each pair of symbols $\{a_i, b_i\}$, a separate SC is required.
\\
\begin{definition}[Stack Circuit]
\label{SC}
For each pair of $\{a_i,b_i\}$, we construct a corresponding SC. An SC  is a circuit consist of  a brain area $A_i$ and two directed fibers $\overrightarrow{A_iA_{i+1}}$, $\overleftarrow{A_iA_{i+1}}$, which connect the brain areas $A_i$ and $A_{i+1}$.
\\
\end{definition}

\begin{figure}[H]
\centering
    \begin{minipage}[t]{0.7\linewidth}
        \centering
        \includegraphics[width=\textwidth]{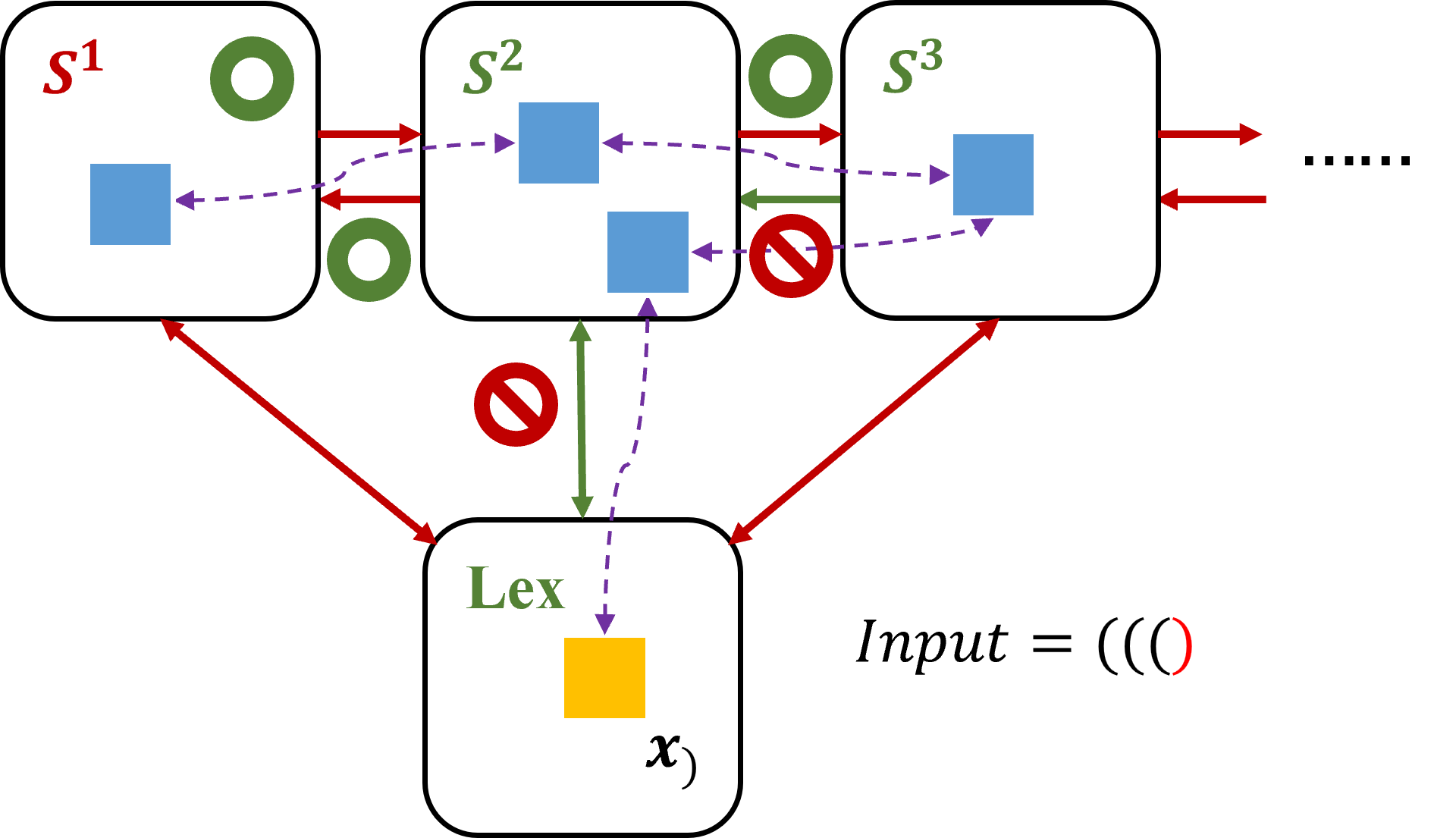}
    \end{minipage}%
    \caption{A sketch of SC, while the input string is $``((()"$.}
    \label{fig:stack}
\end{figure}

Taking the $D_1$ language as an example, whenever a left bracket is input, starting from $S_1$, the index of the next projected area will increase by one. Conversely, when a right bracket is input, the index will decrease by one. Only when the last accepted symbol is ``)'' and the current projected area is $S_1$, the string can be accepted by $D_1$. Following the same principle as the RC, since all fibers are unidirectional, the process of moving from $S_2$ to $S_3$ and back to $S_2$ does not cause confusion between assemblies, as shown in Figure\ref{fig:stack}. 

It should be emphasized that our model depicts the abstract human brain rather than being completely consistent with the real brain. For example, theoretically, the Stack Circuit may require an infinite number of areas. On the contrary, the real human brain not only has a limited number of brain areas, but also it cannot understand an infinitely long sentence. It's actually psychologically difficult for people when there are more than two embeddings, so the center-embedding discussed in this paper is merely a representative instance of CFL phenomena in natural language. 
%

SC can be considered as a variant of ``fallback-again'' operation mentioned in \cite{mitropolsky-etal-2022-center}, while actually there are many problems with this operation - for example, it will face the situation when multiple consecutive brackets in the same direction, which is similar to the $Adj*$ case of RC, those brackets will not be distinguished. 

\section{Discussion And Proof}
\label{sec:4}

In this section, we want to discuss how to transform the BNLP design presented earlier into an automaton, referred to as the Parser Automaton (PA). By clearly defining the construction rules and transition functions of PA, we can formally explore its expressive capabilities. In addition, we will demonstrate that BNLP with RC can be transformed into a PA capable of expressing all RLs; By combining the expression capabilities of stack circuit for a Dyck languages and the CS theorem, a PA will acquire the ability to express all CFLs. Consequently, it can be proven that PA encompasses pushdown automata (PDA).

\subsection{Simplified model}
\label{sec:sm}
Natural language proceeded by human brain contains a great diversity, but only when it is simplified to a certain sub-model can it be formally discussed. Thus in the first stage of research, the language is just considered as {\bf regular languages}, described by regular expressions ({\bf RE}). As is known to all, finite automata (including DFA and NFA) describe syntactic categories, and their description ability is equivalent to that of RE. This section explicates that BNLP with RC can be transformed into a kind of automaton, which is called the Parser Automaton ({\bf PA}), which has the ability of description no less than that of FA. That means, for any RE, a corresponding PA can be constructed according to some rules. Accordingly, for any FA, a PA that accepts the same language can also be constructed in the same way. 

The definition of regular language will not be explained here. Readers can refer to any formal language and automaton textbook. The key element of RE is that it is a closed expression under {\itshape concatenation}, {\itshape union} and {\itshape Kleene closure}, which can be used to recursively generate all RE based on the simplest RE (a single symbol or $\epsilon$, an empty symbol). 

What also needs to be explained here is some unconventional statements to match the characteristics of BNLP and natural languages. In this paper, many part-of-speech names such as Noun, Adj and Verb are used. Actually, they are not symbols in RE, but represent any word of some parts-of-speech. Thus {\itshape variable}, which is a concept in {\itshape Context-Free Grammar} (CFG) is adopted here to express natural languages more conveniently.
\\
\begin{definition}[Variable] \label{def1}
\

A {\itshape variable} is a token representing RE which only contains union of symbols.

Semantically, a variable is corresponded to a certain area in BNLP, representing a
set of symbols which represent words with the syntactic category accepted by the area. Take the RE $(Adj)^*\ N$ as an example:
\begin{equation*}
    Adj=(a_0+a_1+a_2+,\cdots),a_i\ is\ an\ adjective\\
\end{equation*}
\begin{equation*}
    N=(n_0+n_1+n_2+,\cdots),n_i\ is\ a\ noun
\end{equation*}
\end{definition}

Actually, RE discussed in this paper are all augmented RE based on variables rather than symbols. But it doesn't hinder the universality of all conclusions in this paper---because in extreme cases, a variable can be composed of only one symbol. Therefore, a conventional RE can be considered as composed of only variables with a single symbol.

It could be also noticed that there are variables like Sbj and Obj in the parser, which actually both represent Nouns. In fact, they are syntactic categories rather than parts of speech. A word belonging to a certain part of the speech can be accepted by many different STG areas, which is very similar to Nondeterministic Finite Automata (NFA). However, whether a varaible can be accepted by a certain area also depends on the post-rules that have taken effect earlier, e.g., for a RE, $S V O$, both $S,O$ accept nouns and $V$ accepts verbs, when input is a noun, only when there has already been a verb, the noun can be accepted by $O$. In other words, the state to which a state can be transferred is limited by previous fiber-rules, which are determined by the brain areas involved in previous projections. Thus, a coding system in which former input symbols can be reflected in the code of the state is applied, as shown in definition \ref{def:state}. 

For a given parser's parsing rule, it should be expressed in the form of RE\footnote{If there is a rule that RE cannot express, it can only show that the ability of expression of parser is stronger than that of FA, which is not inconsistent with the conclusion of this section.} with no duplicate variables\footnote{That means, for variables such as $S$ and $O$ that accept nouns, as long as they appear in any position of the RE once, it is necessary to give the incoming components a new variable, e.g., a sentence that accepts subject, object, and indirect object, may have $S$, $O$, $I$ three variables which accept nouns. If multiple nominal elements can be modified by adjectives, we might as well define these adjective variables as $Adj^1,\ Adj^2 \cdots$ Each variable actually corresponds to a non-lexicon brain area in the parser, therefore, unlike OP, BNLP will not associate adjectives that modify different nouns with only one brain area, but will establish multiple adjective brain areas.}. And then, the states of that automaton can be coded according to the following rules:
\\
\begin{definition}[State code] \label{def:state}
\

    \begin{enumerate}
    \item Each variable in RE, or a brain area in Parser's brain, corresponds to a digit. Especially for the variables in the Kleene closure, for each Kleene star, it is necessary to set a separate digit to record the number of repetition of the formula in the closure, which is called counting digit (CD).
    \item For a state, if an assembly has been formed on a non-Lex brain area, mark the corresponding digit of the area as 1.
    \item In particular, for the variables in the Kleene closure, if all their corresponding brain areas already have an assembly, then the value of the corresponding counting digit of the closure will be $+1$; Next, the subject to observe by the counting digit will be changed to the recurrent circuit. Every time the recurrent circuit completes a cycle, the counting digit will be $+1$.
\end{enumerate}

\end{definition}

For example, for such a RE, $S\ V\ (O+\epsilon)\ (Prep\ I)^*$, then six digits should be used to code the states, and the starting state should be recorded as $E_{000000}$, acceptance status is $F$, the transition function $\delta$ can be written as:

\begin{equation*}
\begin{aligned}
\delta(E_{000000},n_i )=E_{100000},n_i\ is\ a\ noun
\end{aligned}
\end{equation*}
\begin{equation*}
\begin{aligned}
\delta(E_{100000},v_i )=E_{110000},v_i\ is\ a\ verb
\end{aligned}
\end{equation*}
\begin{equation*}
\begin{aligned}
\delta(E_{110000},n_i )=E_{111000},n_i\ is\ a\ noun
\end{aligned}
\end{equation*}
\begin{equation*}
\begin{aligned}
\delta(E_{11jk00},p_i )=E_{11jk10},p_i\ is\ a\ prep
\end{aligned}
\end{equation*}
\begin{equation*}
\begin{aligned}
\delta(E_{11jk10},n_i )=E_{11jk11},n_i\ is\ a\ noun
\end{aligned}
\end{equation*}
\begin{equation*}
\begin{aligned}
\delta(E_{11jk11},\epsilon )=E_{11j(k+1)00}
\end{aligned}
\end{equation*}
\begin{equation*}
\begin{aligned}
\delta(E_{11jk00},\epsilon)=F,while\ j\in\{0,1\}, k\in\mathbb{N}
\end{aligned}
\label{eq:eg1}
\end{equation*}

\begin{figure*}[!ht] 
  \centering
  \begin{minipage}[b]{0.9\linewidth}
  \subfigure[step 1-2]{
    \begin{minipage}[b]{0.2\linewidth}
      \centering
      \includegraphics[width=\linewidth]{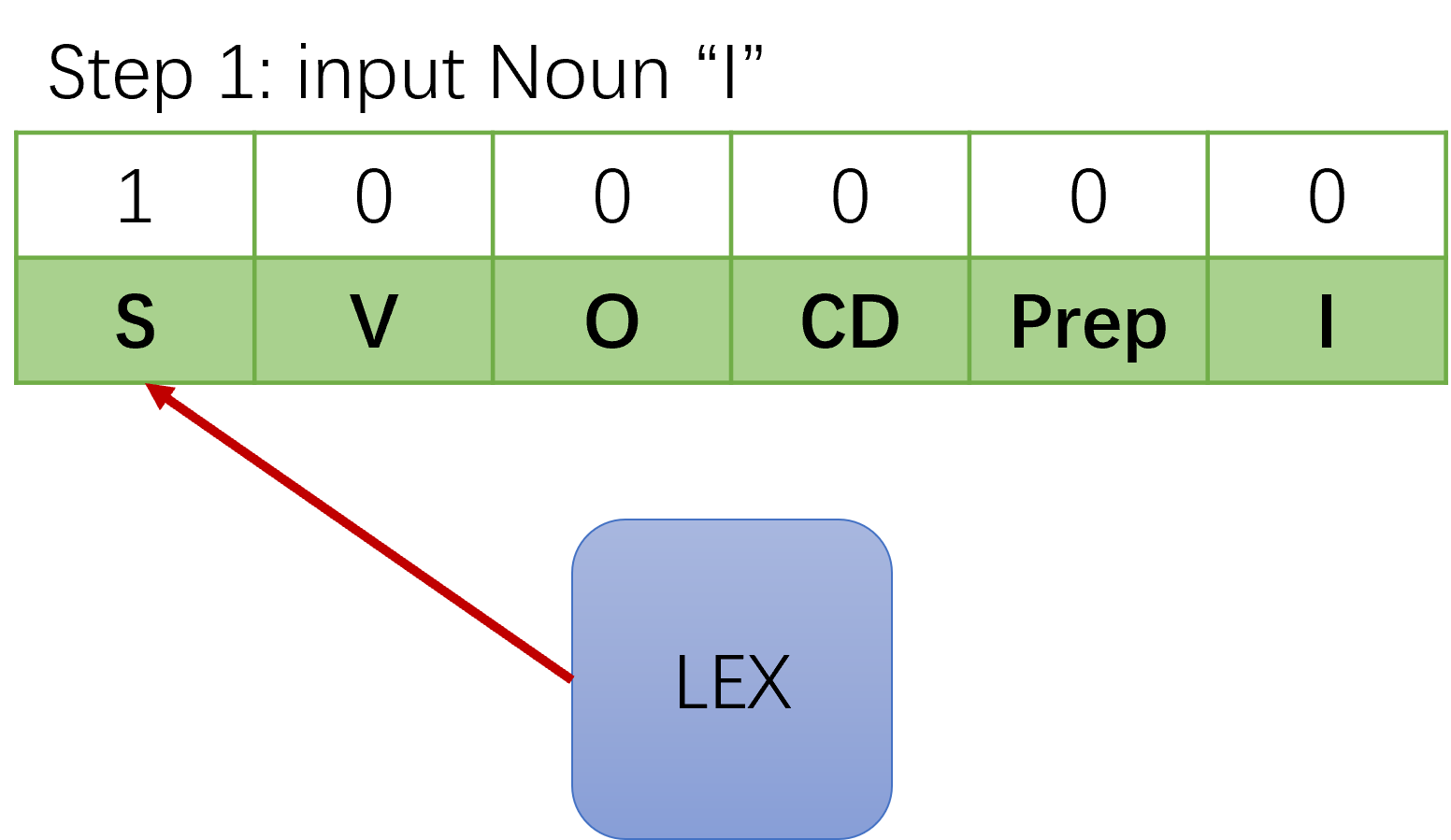}\vspace{8pt}
      \includegraphics[width=\linewidth]{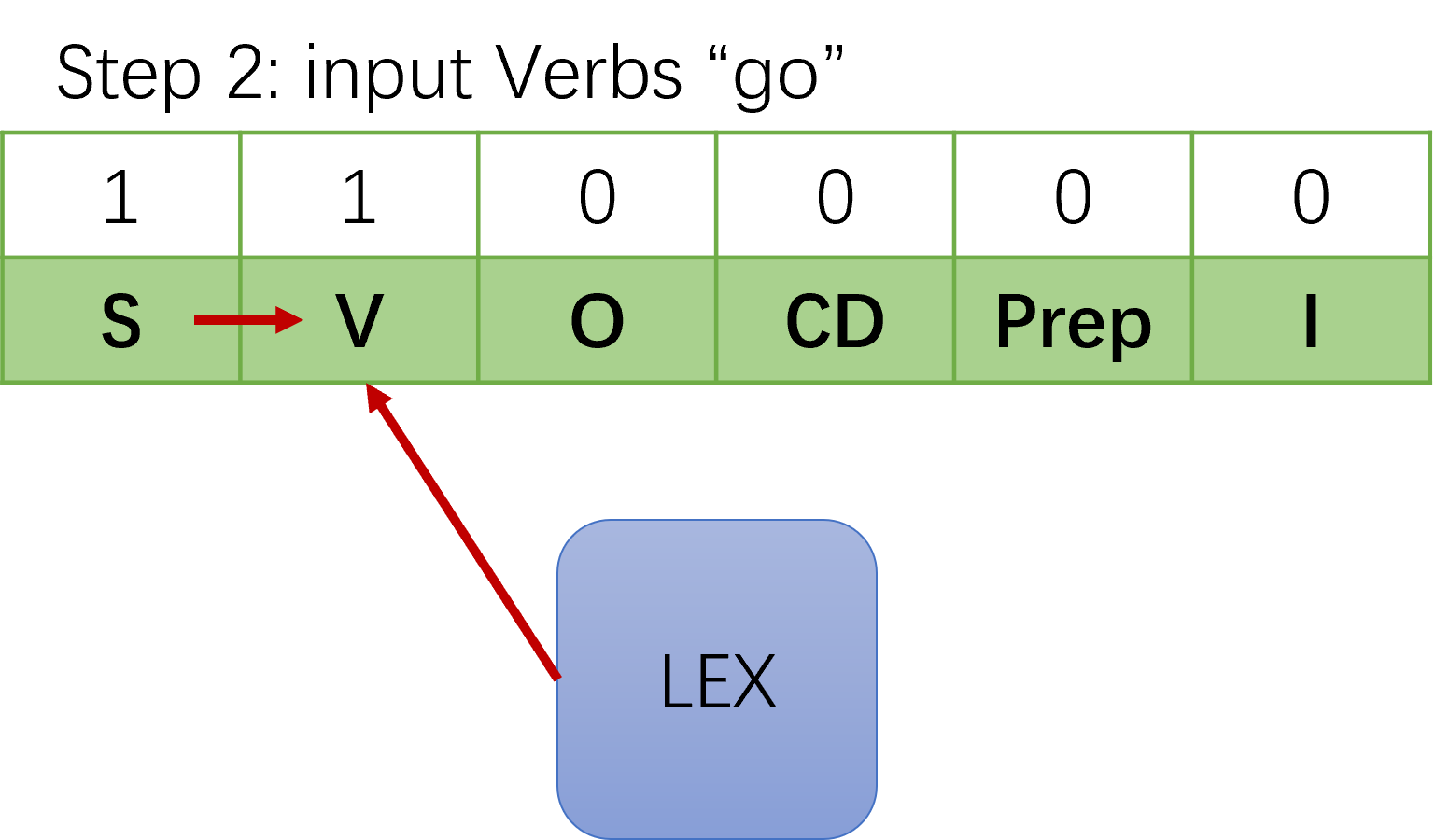}
    \end{minipage}
  }
  \hfill
   \subfigure[step 3-4]{
    \begin{minipage}[b]{0.2\linewidth}
      \centering
      \includegraphics[width=\linewidth]{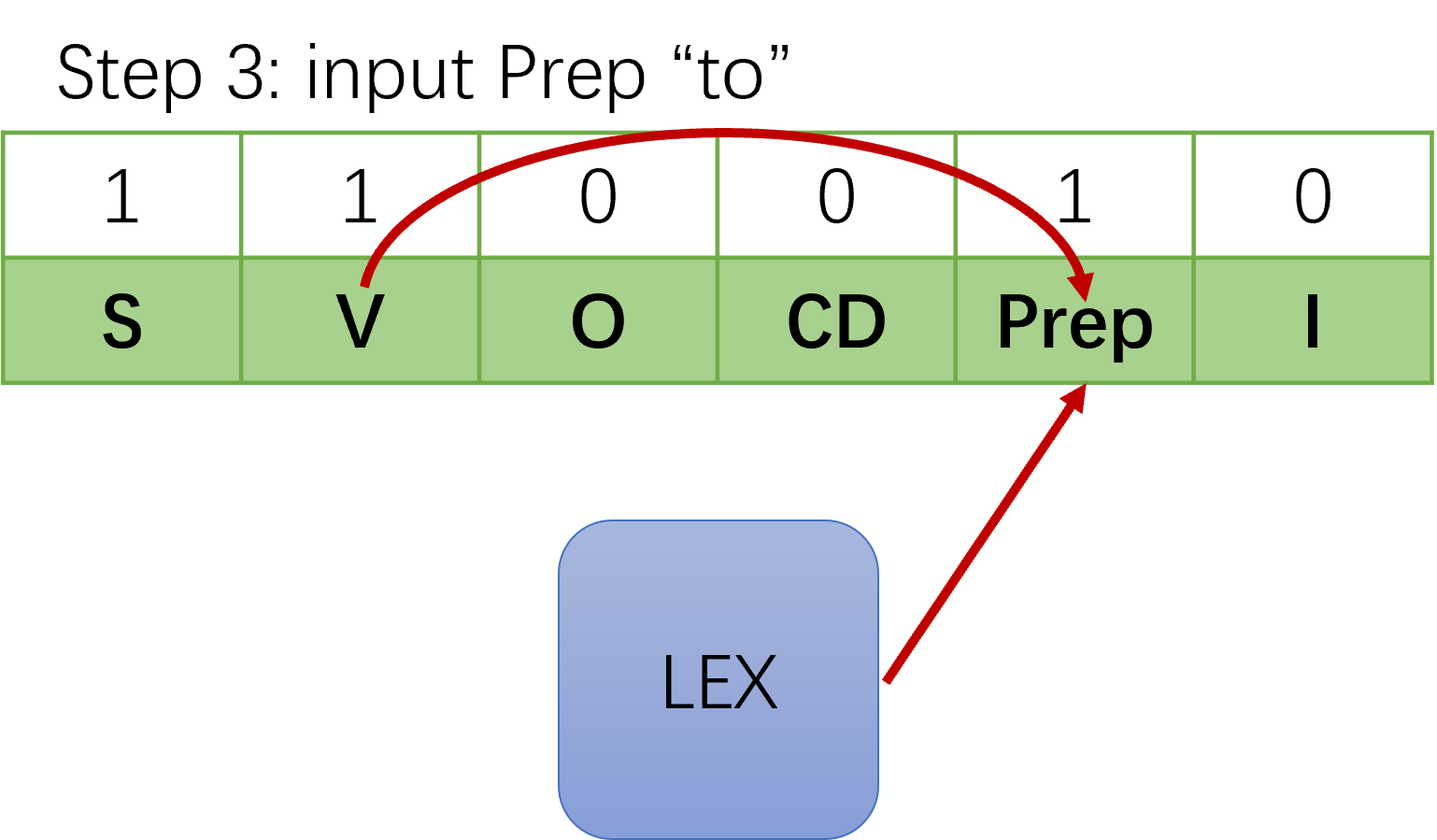}\vspace{8pt}
      \includegraphics[width=\linewidth]{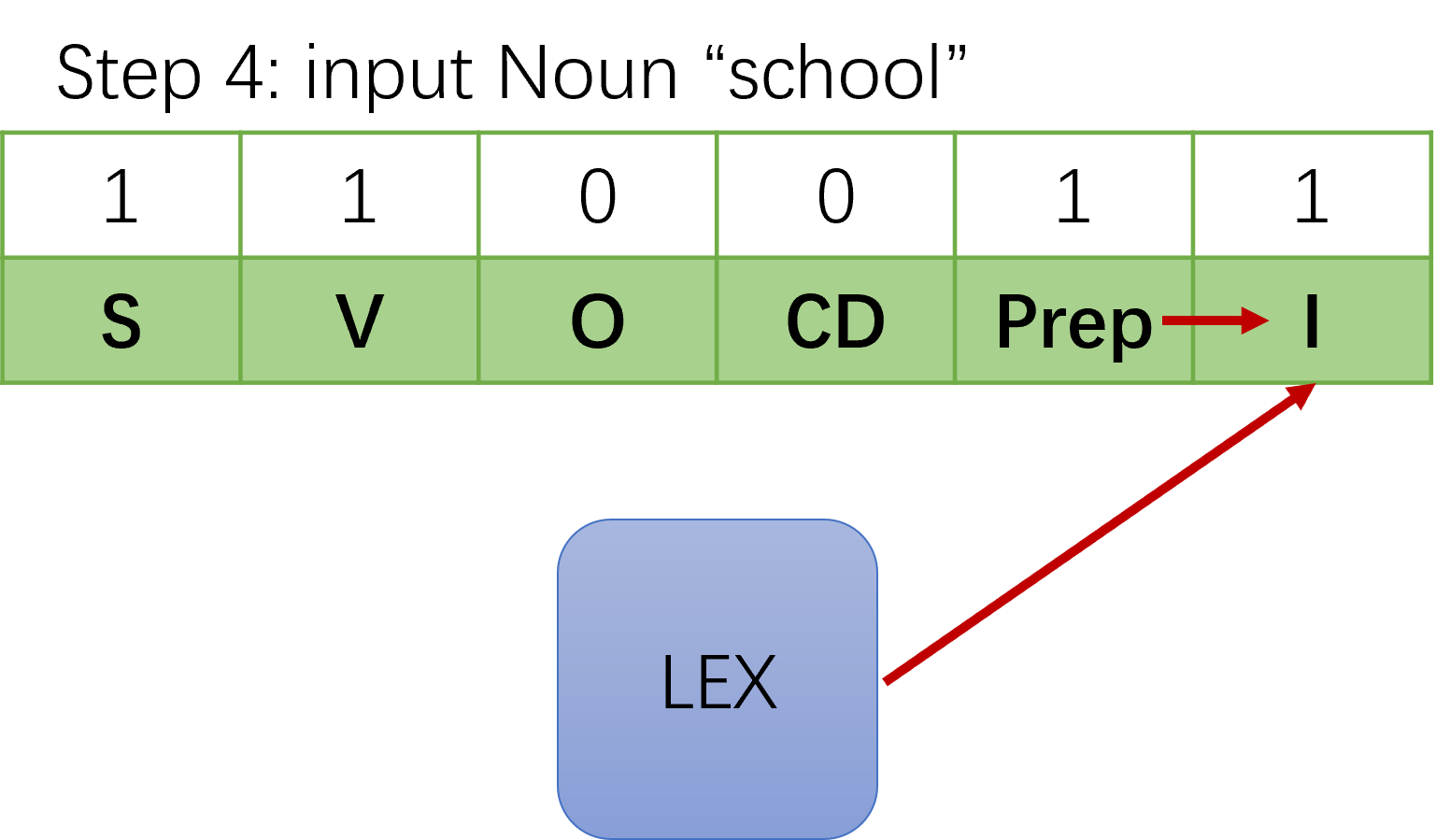}
    \end{minipage}
  }
  \hfill
    \subfigure[step 5-6]{
    \begin{minipage}[b]{0.2\linewidth}
      \centering
      \includegraphics[width=\linewidth]{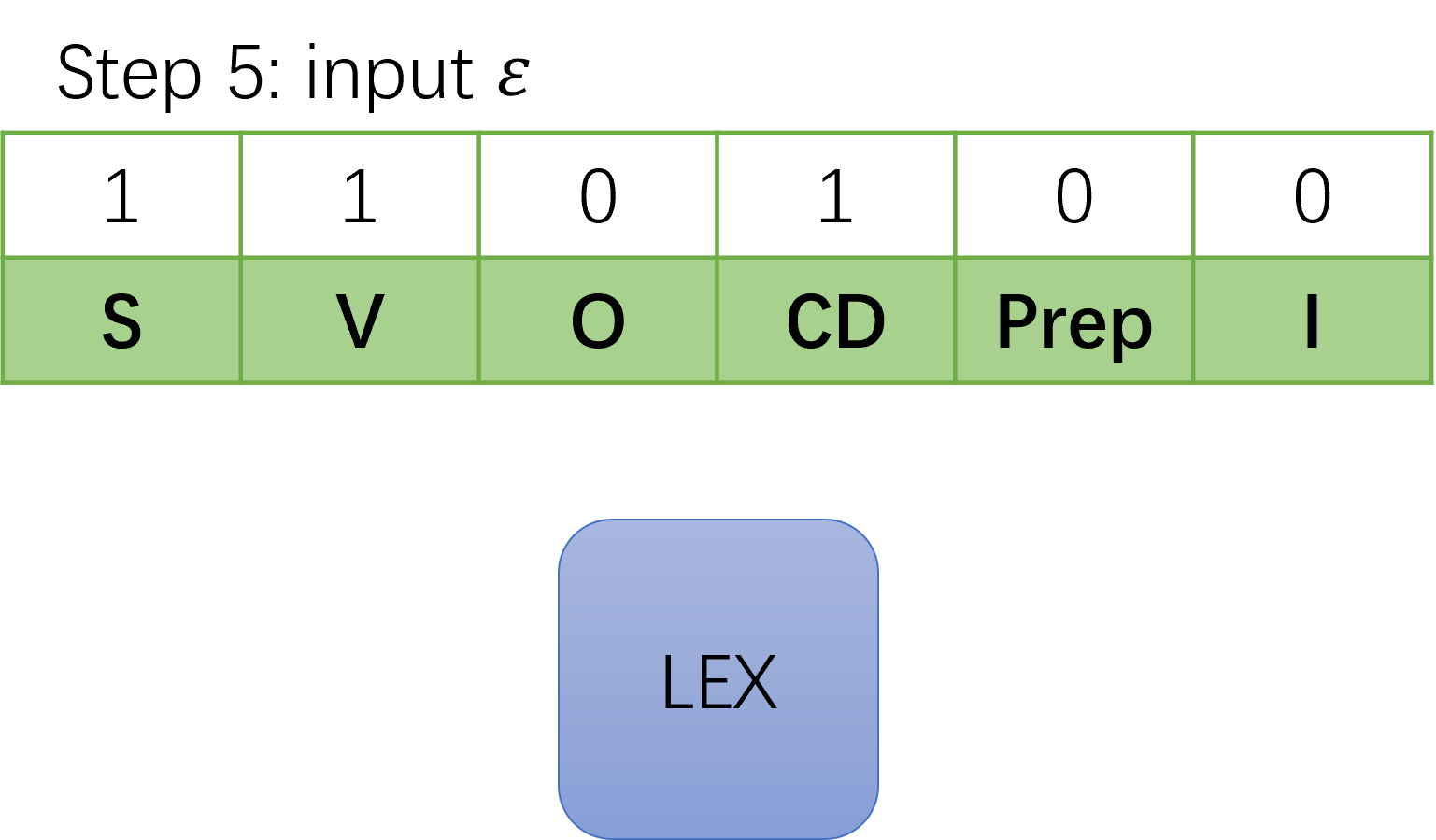}\vspace{8pt}
      \includegraphics[width=\linewidth]{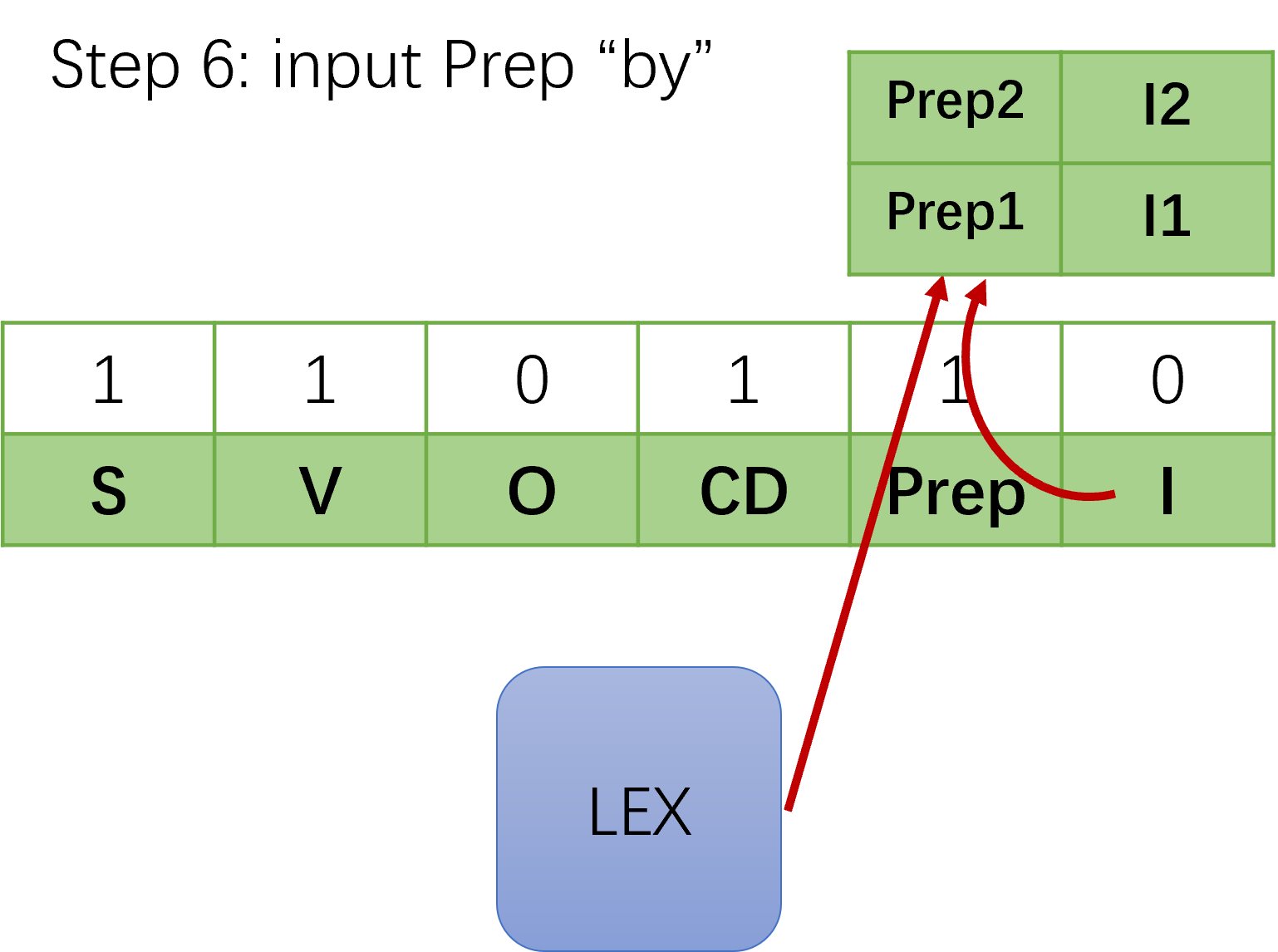}
    \end{minipage}
  }
  \hfill
    \subfigure[step 7-8]{
    \begin{minipage}[b]{0.2\linewidth}
      \centering
      \includegraphics[width=\linewidth]{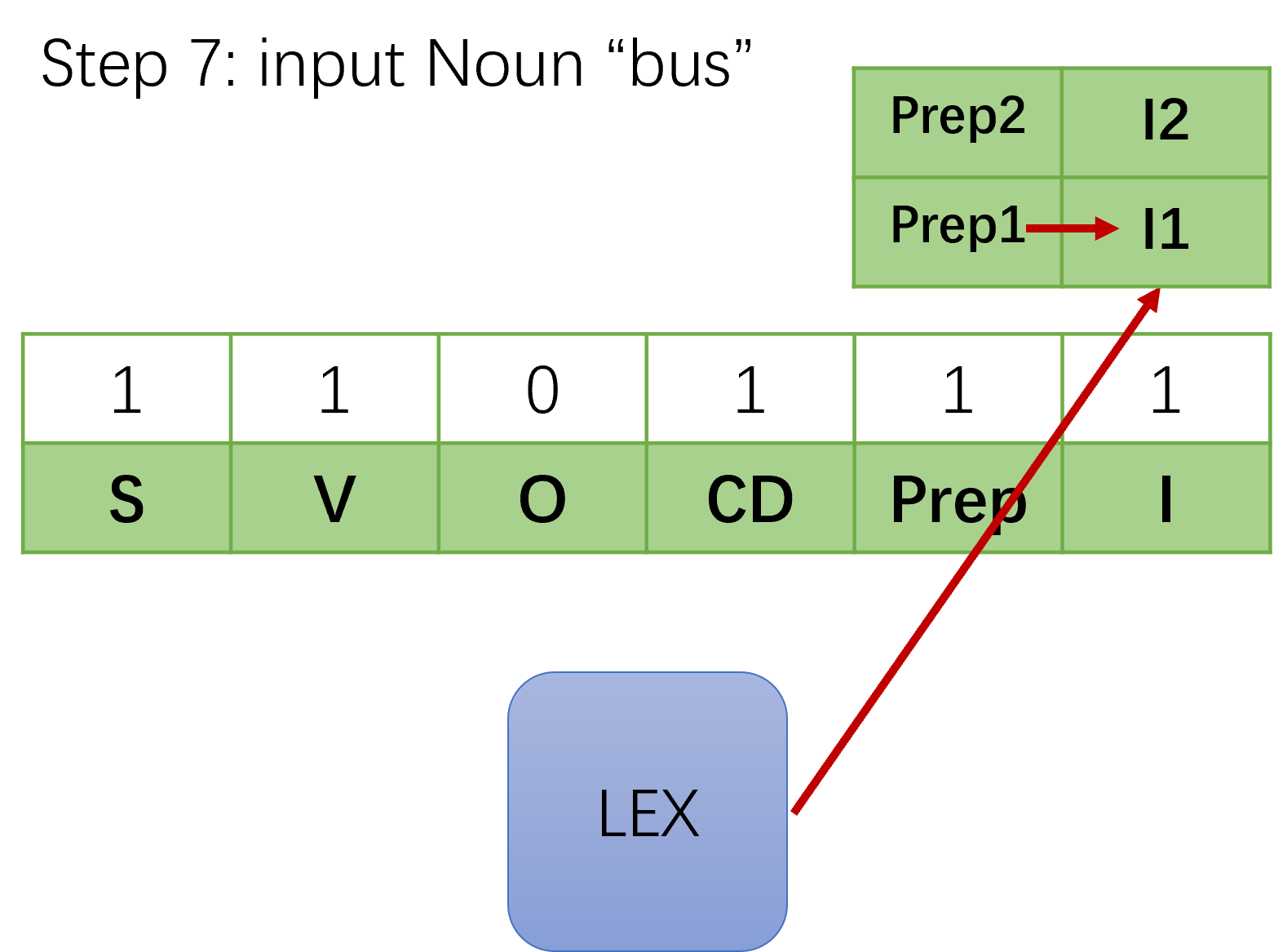}\vspace{8pt}
      \includegraphics[width=\linewidth]{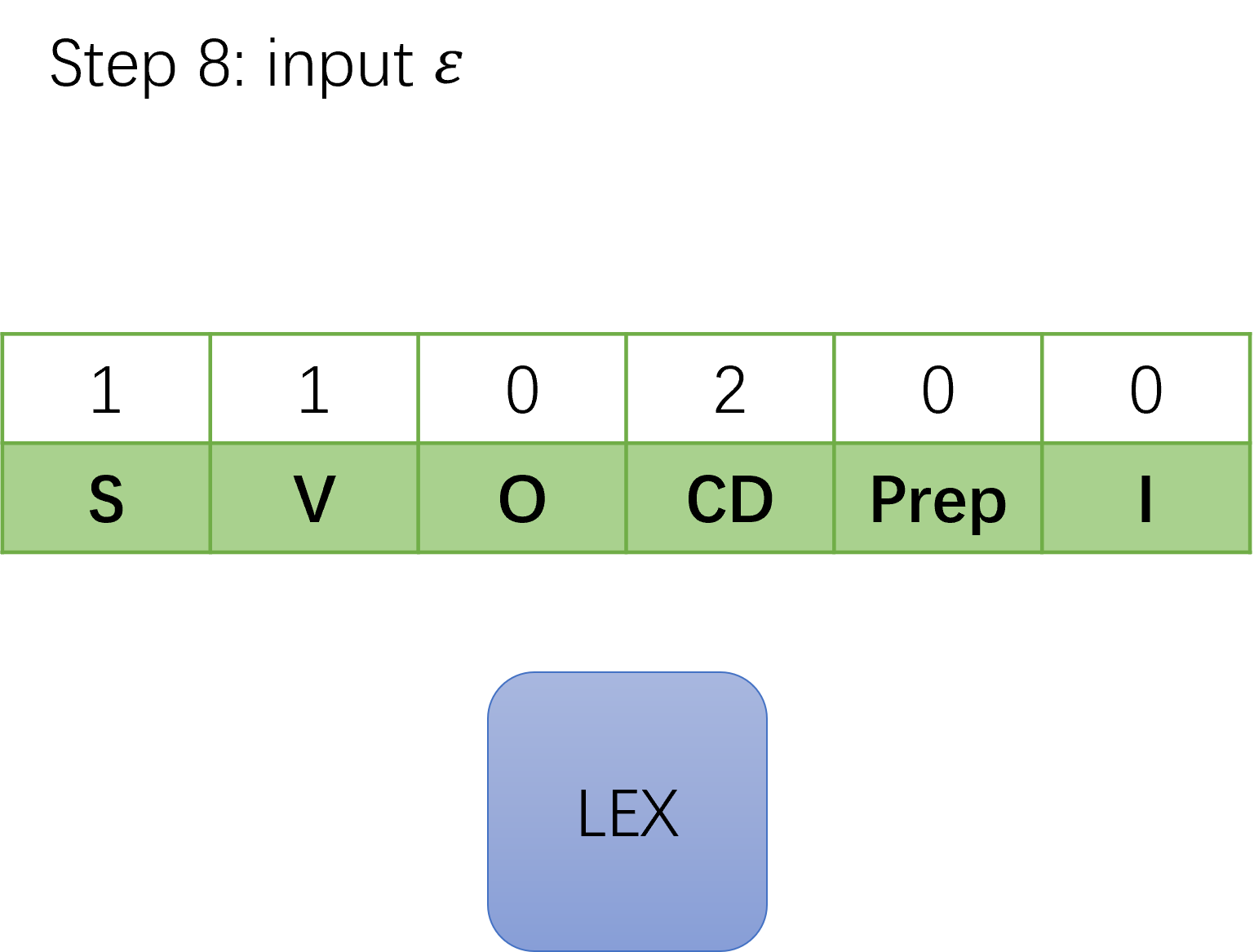}
    \end{minipage}
  }
  \hfill
    \subfigure[step 9]{
    \begin{minipage}[b]{0.2\linewidth}
      \centering
      \includegraphics[width=\linewidth]{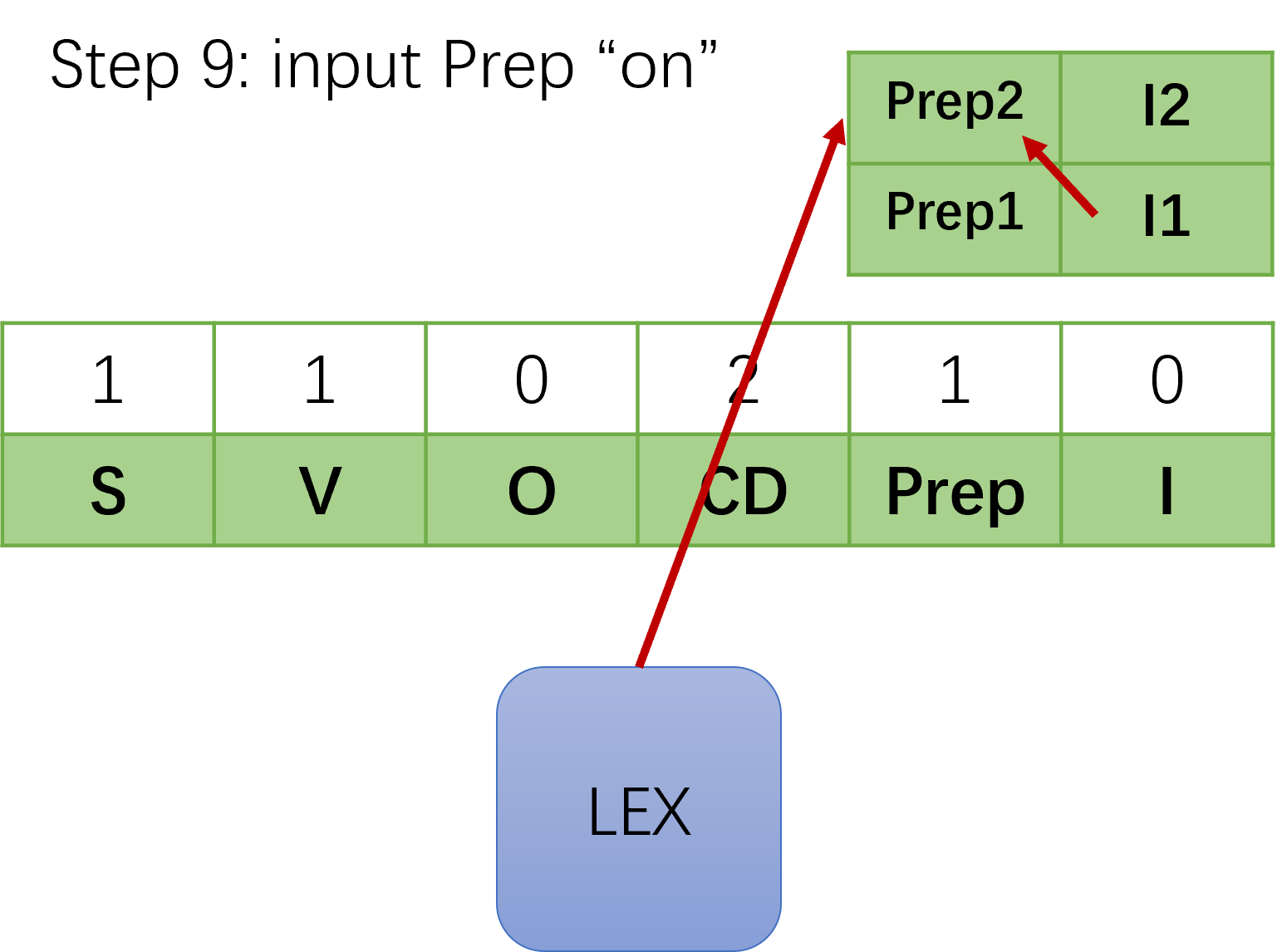}
    \end{minipage}
  }
  \hfill
    \subfigure[step 10]{
    \begin{minipage}[b]{0.2\linewidth}
      \centering
      \includegraphics[width=\linewidth]{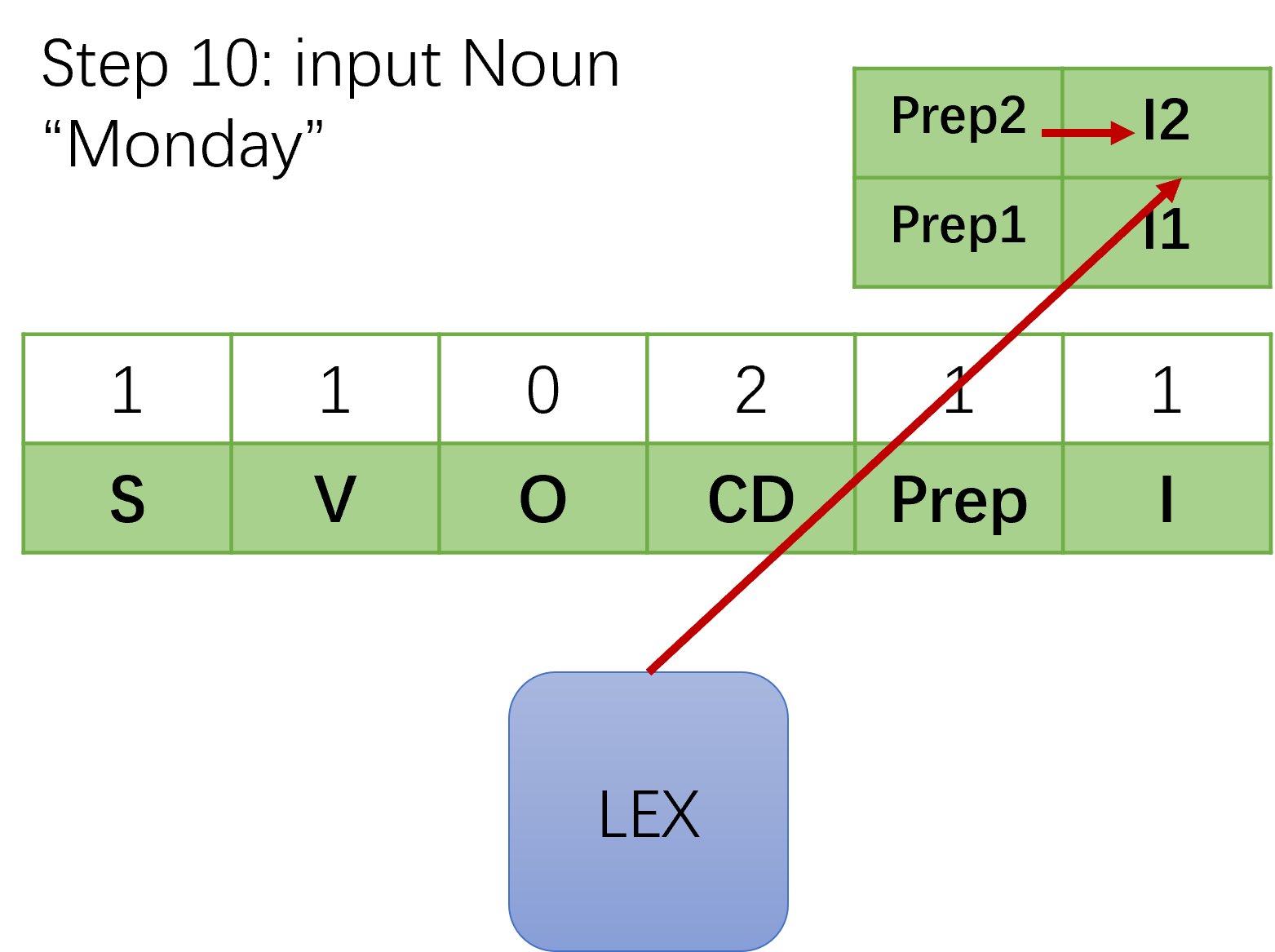}
    \end{minipage}
  }
  \hfill
    \subfigure[step 11]{
    \begin{minipage}[b]{0.4\linewidth}
      \centering
      \includegraphics[width=\linewidth]{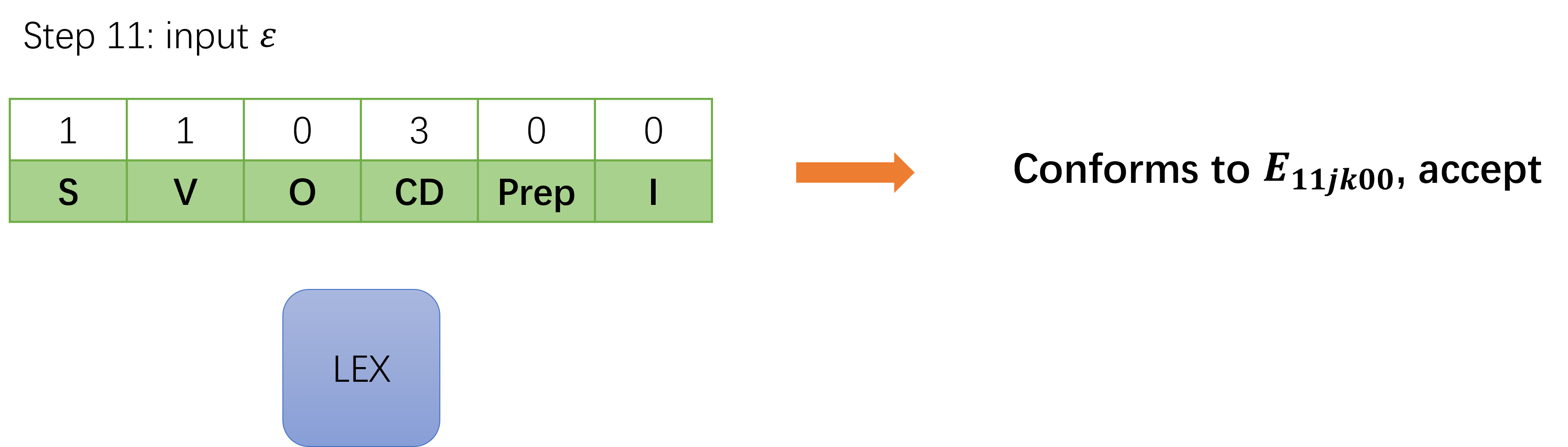}
    \end{minipage}
  }
  \end{minipage}
  \vfill
  \caption{The state transition diagram of the PA of $S\ V\ (O+\epsilon)\ (Prep\ I)^*$, while input ``I go to school by bus on Monday''. The green grids show the states of brain areas, and the white grids record the values of the corresponding digits in the state code, red lines represent the projection relationship among brain areas.}
\label{fig:solong}
\end{figure*}

The detail of how this PA works and how the state codes change can refer to Figure \ref{fig:solong}.

Hence we have got how to generate a PA corresponding to a certain BNLP, and in the next section it will be strictly proved that a PA can always be constructed for any given RE.

\subsection{The closure properties of the parser automaton}
What is needed to prove is that for any RE, a corresponding PA which accepts the language of the RE can be produced. The idea is, is to design a formal system to recursively construct the corresponding PA from the generation rules of RE, and then prove that this system is closed to all RE. So the corresponding PA of the simplest RE (single variable)\footnote{Empty RE ($\varnothing$) is not considered in the following, for such case is trivial. } and the generation rules of PA should be descripted first.
\\
\begin{definition}[PA of the simplest RE]
\label{def2}
\

For the simplest RE, i.e., a single variable $A$, the corresponding PA is easily obtained as:

$P(A) = \{ S_A,\Sigma,\delta_A,E_0,F\}$, where $S_A$\footnote{This section records the transition function of a RE $R$ as $\delta_R$ and its states set as $S_R$.} is the set of states, $\Sigma$ is the input alphabet, $\delta_A$ is the transition function, $E_0$ is the start state and $F$ is the final state. In addition, the alphabet $\Sigma$ is the vocabulary of the natural language to be processed, which is considered the same here.

In that way, there are only three states in $S_A$, i.e., $S_A=\{ E_0,E_1,F\}$. And the transition function $\delta_A$ subjects to:

\begin{equation*}
\begin{aligned}
\delta(E_{0},n_i )=E_{1},n_i\ is\ accepted\ by\ A
\end{aligned}
\end{equation*}
\begin{equation*}
\begin{aligned}
\delta(E_{1},\epsilon )=F
\end{aligned}
\end{equation*}
$\square$
\end{definition}

Although RE discussed below is based on variables rather than symbols, a symbol can be regarded as a variable that accepts only one symbol, so it does not affect the validity of the conclusion. 

Then, the generation rules of PA of new RE generated from existing REs can be summarized as follows.
\\
\begin{definition}[PA generation rules]
\label{def3}
\
    \begin{enumerate}
    \item Concatenation

For RE $A$ and $B$, a new RE $AB$ can be generated by concatenating them. If there are $k_1$ bits in states of $A$, and $k_2$ bits in states of $B$, then for $E_{\overline{n}}\in S_A$, should put the corresponding state $E_{\tiny \overline{n}\ \overline{\underbrace{00\cdots0}_{k_2} }}$ into $S_{AB}$.

Then the transition function $\delta_{AB}$ should inherit all rules in $\delta_A$ but the names of states should be override like above, namely:

For each rule in $\delta_A$, i.e., $\delta_A(E_{\overline{n}},w)=E_{\overline{n'}}$, $E_{\overline{n'}}\neq F$, should be translated as 

\begin{equation}
    \centering
    \delta_{AB}(E_{\tiny \overline{n}\ \overline{\underbrace{00\cdots0}_{k_2} }},w)=E_{\tiny \overline{n'}\ \overline{\underbrace{00\cdots0}_{k_2} }}
\end{equation}

But excepts cases when $E_{\overline{n'}}= F$. For $E_{\overline{m}}\in S_A$, if $\delta_A(E_{\overline{m}},\epsilon)= F$, then for each case in $\delta_{B}$, where $\delta_B(E_{\tiny \overline{\underbrace{00\cdots0}_{k_2} }},w)=E_{\overline{j}}$, add the rule below into $\delta_{AB}$.

\begin{equation}
    \centering
    \delta_{AB}(E_{\tiny \overline{m}\ \overline{\underbrace{00\cdots0}_{k_2} }},w)=E_{\tiny \overline{m}\ \overline{j}}
\end{equation}

Then add states from $S_B$. For $E_{\overline{m}}\in S_A$, if $\delta_A(E_{\overline{m}},\epsilon)= F$, and for each $E_{\overline{n}}\in S_B$, add $E_{\overline{m}\overline{n}}$ into $S_{AB}$.

Then translate rules in $\delta_B$ into $\delta_{AB}$. States in $S_B$ should be renamed by the above rules before added into $S_{AB}$. Then for each rule in $\delta_B$, i.e., $\delta_B(E_{\overline{n}},w)=E_{\overline{n'}}$, should be translated as 

\begin{equation}
    \centering
    \delta_{AB}(E_{\overline{m}\ \overline{n}},w)=E_{\tiny \overline{m}\ \overline{n'}}, \forall{E_{\overline{m}}}\in \{ s\in S_A | \delta_A(s,\epsilon)= F \}
\end{equation}

\item Union

For RE $A$ and $B$, a new RE $A+B$ can be generated by their union. The first step is the same as that of concatenation, namely, renaming the states in $S_A$, as well as treat states in $S_B$ in the same way.  If there are $k_1$ bits in states of $A$, and $k_2$ bits in states of $B$, then for $E_{\overline{n}}\in S_A$, should put the corresponding state $E_{\tiny \overline{n}\ \overline{\underbrace{00\cdots0}_{k_2} }}$ into $S_{A+B}$. Meanwhile, for $E_{\overline{n}}\in S_B$, also put the corresponding state $E_{\tiny \overline{\underbrace{00\cdots0}_{k_1}}\ \overline{n}}$ into $S_{A+B}$.

Then, it is easier than the case of concatenation to translate $\delta_A$,$\delta_B$ into $\delta_{A+B}$.

For each rule in $\delta_A$, i.e., $\delta_A(E_{\overline{n}},w)=E_{\overline{n'}}$, should be translated as

\begin{equation}
    \centering
    \delta_{A+B}(E_{\tiny \overline{n}\ \overline{\underbrace{00\cdots0}_{k_2} }},w)=E_{\tiny \overline{n'}\ \overline{\underbrace{00\cdots0}_{k_2} }}
\end{equation}

For each rule in $\delta_B$, i.e., $\delta_B(E_{\overline{n}},w)=E_{\overline{n'}}$, should be translated as

\begin{equation}
    \centering
    \delta_{A+B}(E_{\tiny \overline{\underbrace{00\cdots0}_{k_1} }\ \overline{n}},w)=E_{\tiny \overline{\underbrace{00\cdots0}_{k_1} }\ \overline{n'}}
\end{equation}

\item Kleene star

For RE $A$, its Kleene closure can be expressed as $(A)^*$, which is also an RE. Its states set $S_{(A)^*}$ and transition function $\delta_{(A)^*}$ can be relatively easily calculated. For $\forall{E_{\overline{n}}\in S_A}$, an infinity of states $E_{j\overline{n}},j \in \mathbb{N}$ should be added into $S_{(A)^*}$.

Then produce $\delta_{(A)^*}$. For each rule in $\delta_A$, $\delta_A(E_{\overline{n}},w)=E_{\overline{n'}}$, an infinity of rules should be added into  $\delta_{(A)^*}$ in the same way as states, namely,
\begin{equation}
    \centering
    \delta_{(A)^*}(E_{j\overline{n}},w)=E_{j\overline{n'}},j\in \mathbb{N}.
\end{equation}

But also, some additional rules should be concluded:

For $\delta_A(E_{\overline{n}},w)=F$,
\begin{equation}
    \centering
    \delta_{(A)^*}(E_{j\overline{n}},\epsilon)=E_{\tiny (j+1)\overline{\underbrace{00\cdots0}_{k}}}
\end{equation}

while there are k digits in states of A.

This rule is named as ``{\bf the recurrent rule}''.
\\
\end{enumerate}
\end{definition}
\par
Under these definitions, it can be easily proved by mathematical induction that for all RE a corresponding PA which accept the language of the RE is constructible.
\begin{thm}
\label{thm1}
For any RE, a parser automaton can be constructed which accepts the language of that RE.
\end{thm}

\begin{proof}[Poof of Theorem \ref{thm1}]
\

Mathematical induction will be used to strictly prove the theorem.

{\bf Basic 1:} if $A$ is an one-variable RE, $P(A)$ is constructible which accepts $L(A)$ according to definition \ref{def2}.

{\bf Basic 2:} $\epsilon$ is a RE, whose corresponding PA is $P(\epsilon)=\{\{F\},\Sigma,\delta_\epsilon,F,F \}$, in which $\delta_\epsilon(F,\epsilon)=F$.

{\bf Induction 1:} If $R_1,R_2$ are RE, then $R_1R_2$ is a RE, whose corresponding PA is $P(R_1R_2)$, which is defined according to Concatenation rule in definition \ref{def3}. Due to $L(R_1R_2)=L(R_1)\circ L(R_2)$, we have $s=m\circ n, \forall s\in L(R_1R_2)$, where $m\in L(R_1),n\in L(R_2)$. $L(R_i)$ is accepted by $P(R_i)$, so we have
$$
    \delta_{R_1}(E_{\overline{m}},\epsilon)=F,
    \delta_{R_2}(E_{\overline{n}},\epsilon)=F
$$
According to definition, we have
\begin{equation*}
\begin{split}
&\delta_{R_1R_2}(E_{\overline{m}\overline{n}},\epsilon)=F,\\& \forall E_{\overline{m}}\in \{s\in S_{R_1}|\delta_{R_1}(s,\epsilon)=F\},\\ &\forall E_{\overline{n}}\in \{s\in S_{R_2}|\delta_{R_2}(s,\epsilon)=F\},
\end{split}
\end{equation*}
thus $\delta_{R_1R_2}(E_{\overline{m}\overline{n}},\epsilon)=F$, $s=m\circ n$ is accepted by $P(R_1R_2)$, that is, $P(R_1R_2)$ accepts $L(R_1R_2)$.

{\bf Induction 2:} If $R_1,R_2$ are RE, then $R_1+R_2$ is a RE, whose corresponding PA is $P(R_1+R_2)$, which is defined according to Union rule in definition \ref{def3}. Due to $L(R_1+R_2)=L(R_1)\cup L(R_2)$, what we need to concern is if $L(R_1),L(R_2)$ is accepted by $P(R_1+R_2)$. For $\forall s\in L(R_1)$, the state $\overline{s}$ when consuming all inputs in s satisfies $\delta_{R_1}(E_{\overline{s}},\epsilon)=F$. According to definition, the state will be rewritten as $E_{\overline{n}\ \overline{\underbrace{00\cdots0}_{k_20s} }}$, while
$$
    \delta_{A+B}\left(E_{\overline{n}\ \overline{\underbrace{00\cdots0}_{k_20s} }},\epsilon\right)=F
$$
Thus $P(R_1+R_2)$ accepts s. Similarly, for $\forall s\in L(R_1)$, $P(R_1+R_2)$ accepts s. In summary, $P(R_1+R_2)$ accepts $L(R_1+R_2)$.

{\bf Induction 3:} If $R_1$ are RE, then $(R_1)^*$ is a RE, whose corresponding PA is $P((R_1)^*)$, which is defined according to Kleene star rule in definition \ref{def3}. For $\forall s\in L((R_1)^*)$, $s=m_1\circ m_2\circ\dots\circ m_n,m_i\in L(R_1)$. For every $m_i$, we have
$$
    \delta_{R_1}(E_{\overline{m_i},\epsilon})=F
$$
then
$$
    \delta_{(R_1)^*}(E_{j\overline{m_i},\epsilon})=
    E_{j\ \overline{\underbrace{00\cdots0}_{k0s} }}
$$
where k denotes the digit count of state code of $R_1$. Thus the final state after consuming $s$ becomes $E_{n\ \overline{\underbrace{00\cdots0}_{k0s} }}$, which satisfies 
$$
\delta_{(R_1)^*}(E_{j\ \overline{\underbrace{00\cdots0}_{k0s} }},\epsilon)=F
$$
for any natural number $j$. Thus $P((R_1)^*)$ accepts s, that is, $P((R_1)^*)$ accepts $L((R_1)^*)$

For that the complete set of RE is a closed set under concatenation, union and Kleene star, any RE $R$ can be generated by these three operations, and because of above inductions, a corresponding PA $P(R)$ which accepts $L(R)$ is constructible.

\end{proof}

\subsection{The parser automata ${\small \supseteq}$ FA}

Based on theorem \ref{thm1}, it is easy to prove that PA is an automaton with stronger ability of description than that of FA. 

\begin{thm}
    \label{thm2}
    {\bf PA} $\supseteq$ {\bf FA}
\end{thm}

\begin{proof}[Poof of Theorem \ref{thm2}]

    For every FA, a RE defining its language can be found, such proof can be found in any Formal Language and Automata textbook. And theorem \ref{thm1} shows that for every RE, a PA which accepts its language can be constructed according to rules in definition \ref{def3}. Thus, for every FA, a PA which can parse its language is constructible. 
\end{proof}

\subsection{Stack circuit's capability to parse Dyck languages}
\label{sec:dyck}

Similar to proving the parsing ability of PA for regular languages, we also need to strictly explain why an automaton with a Stack Circuit can parse Dyck languages. The best approach is to have a similar state encoding mechanism that assigns a unique state encoding to the automaton for each input string and explains how to determine whether the automaton accepts the current string. Based on the previous definition of state code, this task is relatively straightforward.
\\
\begin{definition}[Stack States]
\

    For a Dyck language, $D_k=\{a_1,\cdots,a_k,b_1,\cdots,b_k\}$, each pair of symbols ${a_i, b_i}$ is associated with a digit $d_i$. These digits follow the assignment rules below:
    \begin{enumerate}
        \item When there is no input, all $d_i$ values are $0$;
        \item If the character input at the current state is $a_i$, the corresponding $d_i$ value should be incremented by one;
        \item If the character input at the current state is $b_i$, the corresponding $d_i$ value should be decremented by one.
    \end{enumerate}
\end{definition}
 
Using Figure\ref{fig:stack} as an example again, when the input string is $``((()"$, the current state is $E_2$; if the input string is $``()(())"$, the state is $E_0$.


An automaton accepting Dyck languages can easily be designed based on this definition, i.e., an automaton that only accepts states with all $d_i$ values being $0$. However, one possible concern is whether this design would cause ambiguity, as both strings $``()()"$ and $``((()))"$ correspond to state $E_0$.

For an automaton parsing Dyck languages, this design is sufficient. We need to consider what these brackets, which do not represent natural language vocabulary, actually represent in natural language, in fact, they stand for guiding words like "that" or "which" or even punctuation marks separating clauses, and they do not include any content in between. When combining the Dyck languages representing the center-embedding process with the PA expressing regular languages, $d_i$ will function similarly to counting digits, indicating whether the center-embedding process within the clause is closed. Even for states with identical $d_i$ values, there will be distinct values for other sentence components corresponding to different digits.

It's evident that SC has the capability to parse  Dyck languages, and the construction of the automaton corresponding to SC is straightforward. Indeed, SC constitutes a form of PDA:

\begin{equation}
    \centering
            P(S_{A_i})=\{\{q_0,P\},\{a_i,b_i\},\delta,\{d_0\},q_0,d_0,\{q_0\}\}.
\end{equation}

where $S_{A_i}$ denotes an SC containing a brain area $A_i$,  $q_0$ denotes the start state,  $P$ denotes the state when $d_i\ne0$. When the stack is empty, we have $\delta(P, \epsilon, d_0)=\{q_0,d_0\}$. Input can be accepted if and only if the state is $q_0$ and the stack is empty.

\subsection{The parser automata ${\small \supseteq}$ PDA}

Based on Chomsky-Schützenberger theorem \cite{CHOMSKY1963118, Kaoru2010}, it is to prove that PA with both RC and SC is capable to parser any CFL, and has ability of description no less than that of PDA. Due to PA with SC accepts all Dyck languages (discussed in \ref{sec:dyck}), and \ref{thm2} has shown that PA accepts all RLs. Therefore, for any PDA, the CFL $L$ corresponding to it can be considered in the form of $L=h(R \cap D)$, while both $R$ and $D$ can be accepted by a PA. Logically,  we can demonstrate that PA already possess the capability to parse CFL.  However, some details still should be discussed.

Before giving our proof, we shall first introduce the corollary proposed by Ullman et al. \cite{Hopcroft1969FormalLA}.

\begin{thm}
    \label{cs_ullman}
    A CFL L can be represented as the following form:
$$
    L=\{R,\{a_1,a_2,\dots,a_k,b_1,b_2,\dots,b_k\},P,S\}
$$
where $R$ denotes RL, P denotes production rules, which contains the rule $S\rightarrow a_iSb_i$, for $i=1,\dots, k$
\\
\end{thm}
With the help of the corollary above, we can prove the following theorem:
\begin{thm}
    \label{thm3}
    {\bf PA} $\supseteq$ {\bf PDA}
\end{thm}

\begin{proof}[Poof of Theorem \ref{thm3}]


What we're going to prove is that, given any PDA, the corresponding language, a CFL $L$, we can construct a PA to parse L.

According to \textbf{Theorem} \ref{cs_ullman}, we have 
$$
    L=\{R,\{a_1,a_2,\dots,a_k,b_1,b_2,\dots,b_k\},P,S\}
$$
and  production rules $S\rightarrow a_iSb_i,i=1,\dots,k$. Since any pair of $\{a_i, b_i\}$ possess same properties, we can merely discuss the case with exactly one production rule $S\rightarrow aSb$(we can cover cases with more pairs of $\{a_i, b_i\}$ by adding more states). By regarding every string in $R$ as a terminal, $L$ can be written as the form of CNF:
\begin{equation}
\begin{split}
S &\rightarrow AB \\
C &\rightarrow SB \\
S &\rightarrow SS \\
S &\rightarrow r, \quad r \in R \\
A &\rightarrow a, \quad B \rightarrow b
\end{split}
\end{equation}

the corresponding PDA should be $$P(L)=\{\{q_0,P\},\{a,b,r\},\delta,\{d_0\},q_0,d_0,\{q_0\}\}$$ with a transition function:

\begin{equation}
\begin{split}
    \delta(q_0,a,d_0)&=\{(P, Xd_0)\}     \\
    \delta(q_0,r,d_0)&=\{(q_0, d_0)\}     \\
    \delta(P,a,X)&=\{(P, XX)\}     \\
    \delta(P,r,X)&=\{(P, X)\}     \\
    \delta(P,b,X)&=\{(P, \epsilon)\}     \\
    \delta(P,\epsilon,d_0)&=\{(q_0, d_0)\}     \\
\end{split}
\end{equation}
Obviously, the transition of $P(L)$ corresponds to how SC cope with a string $r$ in an RL, thus we have $SC=P(L)$, that is, PA with SC can parse $L$, which means $PA \supseteq PDA$.
\end{proof}

\section{Summary}
\label{summary}
By now, we have proposed a modified AC parser that includes RC, which can be proved to have stronger description ability than that of FA. Furthermore, inspired by the CS theorem, we demonstrate that we can construct a BNLP with the ability to parse any given CFL by introducing a structure SC that can parse Dyck languages. Thus, the application of AC to Natural Language Processing will gain a theoretical fulcrum. 

According to Papadimitriou et al.'s initial conception \cite{papadimitriou2020brain}, they tried to parse natural language by introducing an option 'Merge' which was inspired by CNF, however, this attempt was eventually abandoned\footnote{We encountered many issues in the experiments concerning Merge and other operations, namely, Hebbian runaway dynamics and representational drift \cite{Jphy22}. Therefore, exploring the principles of computational neuroscience behind these issues and finding corresponding solutions to improve the AC model will be our further research focus.}. Our method provides a path to achieve advanced brain functions based entirely on AC's basic settings (i.e., areas, fibers, and projection maps), avoiding various unpredictable complex system issues that arise when packaging low-level neural projections into advanced operations, while being mathematically reliable.

\section*{Acknowledgments}
The corresponding author of this work is Jianlin Feng. This work is partially supported by China NSFC under Grant No. 61772563.



\bibliographystyle{IEEEtran}


\end{document}